%% file: main.tex
\documentclass[runningheads]{llncs}
\usepackage[T1]{fontenc}
\usepackage{graphicx}
\usepackage{amsmath,amssymb}
\usepackage{xcolor}
\usepackage{adjustbox}
\usepackage{colortbl}
\usepackage{tabularx}
\usepackage{booktabs} 
\usepackage{graphicx}
\usepackage[linesnumbered, vlined, ruled]{algorithm2e}
\SetArgSty{textup}
\usepackage{enumerate}
\usepackage{prettyref}
\usepackage{color, colortbl}
\usepackage{braket}
\definecolor{MyHiLiRow}{gray}{0.9}
\usepackage{tcolorbox}
\usepackage{caption}
\usepackage{seqsplit}
\usepackage{bm}
\usepackage{nameref}
\usepackage{multirow}
\usepackage{xr}
\usepackage{tikz}
\usepackage{enumitem}

\usepackage{subcaption}

\usepackage{mathtools}
\usepackage{enumitem}

\def\vec#1{\mathbf{#1}}
\def\set#1{\mathcal{#1}}

\definecolor{refpt}{RGB}{44,160,44}   
\definecolor{nearpt}{RGB}{255,127,14} 
\definecolor{solpt}{RGB}{31,119,180} 
\definecolor{candpt}{RGB}{128,128,128}
\definecolor{anearpt}{RGB}{255,0,0}
\definecolor{ideal}{RGB}{123, 94, 167}
\definecolor{nadir}{RGB}{140, 81, 10} 

\newcommand{\pref}[1]{\prettyref{#1}}
\newrefformat{fig}{Fig.~\ref{#1}}
\newrefformat{supfig}{Fig.~\ref{#1}}
\newrefformat{tab}{Table~\ref{#1}}
\newrefformat{suptab}{Figure~\ref{#1}}
\newrefformat{sec}{Section~\ref{#1}}
\newrefformat{subsec}{Section~\ref{#1}}
\newrefformat{app}{Appendix~\ref{#1}}
\newrefformat{alg}{Algorithm~\ref{#1}}
\newrefformat{property}{Property~\ref{#1}}
\newrefformat{theorem}{Theorem~\ref{#1}}
\newrefformat{corollary}{Corollary~\ref{#1}}
\newrefformat{proposition}{Proposition~\ref{#1}}
\newrefformat{def}{Definition~\ref{#1}}
\newrefformat{eq}{equation~(\ref{#1})}

\begin{document}
%
\title{Effects of Objective Normalization on Regions of Interest in Preference-Based Evolutionary Multi-Objective Optimization}
\titlerunning{Effects of Objective Normalization on ROIs in PBEMO}
%
\author{Ryuichi Mogami\orcidID{0009-0004-6247-775X} and 
Ryoji Tanabe\orcidID{0000-0003-4049-0393}}
%
%
\institute{Yokohama National University, Yokohama, Japan\\
\email{n.ryuichi.mogami@gmail.com, tanabe-ryoji-sn@ynu.ac.jp}}
\maketitle              

\input{abstract}

\input{introduction}
\input{preliminary}
\input{roi_definition}

\input{setup}

\input{results}
\input{conclusion}


\subsubsection*{Acknowledgments.}

This work was supported by JSPS KAKENHI Grant Numbers \seqsplit{25K03194} and \seqsplit{23H00491}.

\bibliographystyle{splncs04}
\bibliography{reference}

\end{document}

%% file: abstract.tex
\begin{abstract}

%

Preference-based evolutionary multi-objective optimization (PBEMO) aims to approximate a region of interest (ROI) defined by the preference information from a decision maker (DM).
Although objective functions in real-world applications typically have different scales, the issue of how to define the ROI in such problems has been overlooked in the literature.
In fact, it has not been standardized in the EMO community whether the ROI should be defined in the unnormalized objective space or in the normalized objective space.
In this context, this paper investigates the effects of objective normalization on ROIs. 
First, this paper shows that two ROIs defined in the unnormalized and normalized objective spaces can differ significantly for problems with differently scaled objectives.
Then, we demonstrate that ROIs defined in the normalized objective space are highly difficult to approximate even on problems with equally scaled objectives because of poor approximations of the ideal and nadir points.
In contrast, we show that ROIs defined in the unnormalized objective space are much easier to approximate than those defined in the normalized objective space.


\keywords{Preference-based evolutionary multi-objective optimization  \and region of interest \and objective normalization}
\end{abstract}

%% file: introduction.tex
\section{Introduction}
\label{sec:introduction}


Multi-objective optimization aims to minimize $m$ conflicting objective functions $f_1, \dots, f_m$ simultaneously.
Evolutionary multi-objective optimization (EMO) has been widely used to find a solution set $\set{P}$ that approximates the Pareto front (PF) in the objective space~\cite{Deb01}.
A decision maker (DM) selects the most preferred solution from $\set{P}$ obtained by an EMO algorithm such as NSGA-II~\cite{DebAPM02} in an a posteriori manner.


The DM's preference information is available a priori in some real-world applications (e.g.,~\cite{BurkotovaPKM23,GranadoSSHF25,TangLDWZF20}).
In such cases, it is rational to exploit the preference information provided by the DM during the search.
Unlike EMO, PBEMO aims to approximate a region of interest (ROI)~\cite{PurshouseDMMW14,RuizSL15}, a subset of the PF defined by the DM's preference information.
The ROI is expected to contain only solutions preferred by the DM in the objective space.
Since the DM does not need to examine irrelevant solutions in the decision-making process, their workload can be reduced.
Furthermore, approximating the ROI is generally easier than approximating the whole PF.
This is especially true as the number of objectives $m$ increases~\cite{LiLDMY20}.

This paper considers preference-based optimization using a reference point $\vec{z} \in \mathbb{R}^m$. 
The reference point $\vec{z}$ is provided by the DM, and each element of $\vec{z}$ represents a preferred objective value.
Due to its simplicity, using the reference point $\vec{z}$ is a popular way of expressing the DM's preference~\cite{BechikhKSG15}.

In~\cite{TanabeL24}, it is pointed out that the ``ROI$"$ is a loosely defined term in the EMO community.
To address this issue, the previous study classified ROIs implicitly used in the literature into the following three types: the closest point-based ROI (ROI-C), achievement scalarizing function-based ROI (ROI-A), and Pareto dominance relation-based ROI (ROI-P).
It is also pointed out that existing PBEMO algorithms were implicitly designed to approximate one of the three ROIs.\\


However, the previous study~\cite{TanabeL24} assumes that the scales of the $m$ objectives are the same.
Whereas this is true for some synthetic test problems (e.g., the DTLZ functions~\cite{DebTLZ05}), this is not true for real-world problems~\cite{TanabeI20}.
Since the $m$ objective functions are generally defined independently in real-world applications (e.g., time vs. price), they often have different scales~\cite{HeITWNS21}.

\begin{table}[t]
\centering
\footnotesize
\caption{Types of ROI considered in previous studies.}
\label{tab:type_roi}
\begin{tabular}{llll}
\toprule
Reference & Algorithm & ROI & Space\\
\midrule
\cite{DebS06} & R-NSGA-II & ROI-C & Normalized space\\
\cite{SaidBG10} & r-NSGA-II & ROI-C& Normalized space\\
\cite{ThieleMKL09} & PBEA & ROI-A & Normalized space\\
\cite{LuqueSHCC09} & g-NSGA-II & ROI-P & Unnormalized space\\
\cite{MohammadiOLD14} & R-MEAD2 & ROI-C & Unnormalized space\\
\cite{GoulartC16} & PAR & ROI-P & Normalized space\\
\cite{LiCMY18} & NUMS & ROI-A & Unnormalized space\\
\cite{LiWTJE18} & TMOEA & ROI-A & Unnormalized space\\
\cite{TanabeL24} & Na & ROI-C, ROI-A, ROI-P & Unnormalized space\\
\cite{MogamiT26} & BSF & ROI-C and ROI-P & Unnormalized space\\
\bottomrule
\end{tabular}
\end{table}

For problems with differently scaled objectives, the ROI can be defined in one of the following two spaces:
%
\begin{enumerate}[label=(\arabic*)]
\item the original unnormalized objective space and
\item the normalized objective space obtained using the true ideal and nadir points. 
\end{enumerate}
Unfortunately, there is no consensus in the EMO community on whether the ROI should be defined in the unnormalized objective space or the normalized objective space.
\pref{tab:type_roi} summarizes the types of ROI considered in previous studies.
As shown in \pref{tab:type_roi}, some previous studies define the ROI in the unnormalized objective space, whereas others define it in the normalized objective space.





In this context, this paper investigates how objective normalization affects ROIs.
A previous study~\cite{Tanabe24} examined the effectiveness of normalization methods in PBEMO.
The results in~\cite{Tanabe24} showed that the normalization methods used in PBEMO algorithms perform significantly worse than those used in conventional EMO algorithms, including NSGA-II, in approximating the ideal point, nadir point, and range of the PF.
This is because PBEMO aims to approximate only the ROI and therefore does not attempt to locate the $m$ extreme points.
However, \cite{Tanabe24} focused on objective normalization in PBEMO algorithms, rather than on the normalization of the ROI itself, which is the focus of this paper.


The remainder of this paper is organized as follows.
\pref{sec:preliminary} provides some preliminaries.
\pref{sec:roi_definition} shows that two ROIs defined in spaces (1) and (2) can be totally different.
\pref{sec:setup} describes our experimental setup.
\pref{sec:results} analyzes the performance of R-NSGA-II~\cite{DebS06} with and without objective normalization in approximating the two ROIs.
Finally, \pref{sec:conclusion} concludes this paper.

%% file: preliminary.tex
\section{Preliminary}
\label{sec:preliminary}

\subsection{Multi-objective optimization}
\label{subsec:mo}

Multi-objective optimization aims to find a solution $\mathbf{x} \in \mathbb{X}$ that \textit{minimizes} an objective function vector $\mathbf{f}: \mathbb{X} \rightarrow \mathbb{R}^m$, where $\mathbb{X} \subseteq \mathbb{R}^n$ is the feasible solution space, and $\mathbb{R}^m$ is the objective space. 
Thus, $n$ is the dimension of the solution space, and $m$ is the dimension of the objective space. 

A solution $\mathbf{x}_1$ is said to dominate $\mathbf{x}_2$ if $f_i (\mathbf{x}_1) \leq f_i (\mathbf{x}_2)$ for all $i \in \{1, \ldots, m\}$ and $f_i (\mathbf{x}_1) < f_i (\mathbf{x}_2)$ for at least one index $i$.
We denote $\mathbf{x}_1 \prec \mathbf{x}_2$ when $\mathbf{x}_1$ dominates $\mathbf{x}_2$.
In addition, $\mathbf{x}_1$ is said to weakly dominate $\mathbf{x}_2$ if $f_i (\mathbf{x}_1) \leq f_i (\mathbf{x}_2)$ for all $i \in \{1, \ldots, m\}$.
A solution $\mathbf{x}^\ast$ is a Pareto-optimal solution if $\mathbf{x}^\ast$ is not dominated by any solution in the feasible solution space $\mathbb{X}$.
The set of all Pareto-optimal solutions in $\mathbb{X}$ is called the Pareto-optimal solution set $\mathcal{X}^{*} = \{\mathbf{x}^* \in  \mathbb{X} \,|\, \nexists \mathbf{x} \in  \mathbb{X} \: \text{s.t.} \: \mathbf{x} \prec \mathbf{x}^* \}$.
The image of the Pareto-optimal solution set in the objective space $\mathbb{R}^m$ is called the PF and denoted as $\mathbf{f}(\mathcal{X}^{*})$. 
The ideal point $\vec{p}^\mathrm{ideal}$ and nadir point $\vec{p}^\mathrm{nadir}$ consist of the minimum and maximum values of the PF for $m$ objective functions, respectively.
Formally, for each $i \in \{1, \ldots, m\}$, $p^\mathrm{ideal}_i = \min_{\vec{x}^* \in \set{X}^*} \{f_i(\vec{x}^*)\}$, and $p^\mathrm{nadir}_i = \max_{\vec{x}^* \in \set{X}^*} \{f_i(\vec{x}^*)\}$.


\begin{figure}[t]
\newcommand{\wir}{0.4}
\centering
\subfloat[ROI-C and ROI-A]{\includegraphics[width=\wir\textwidth]{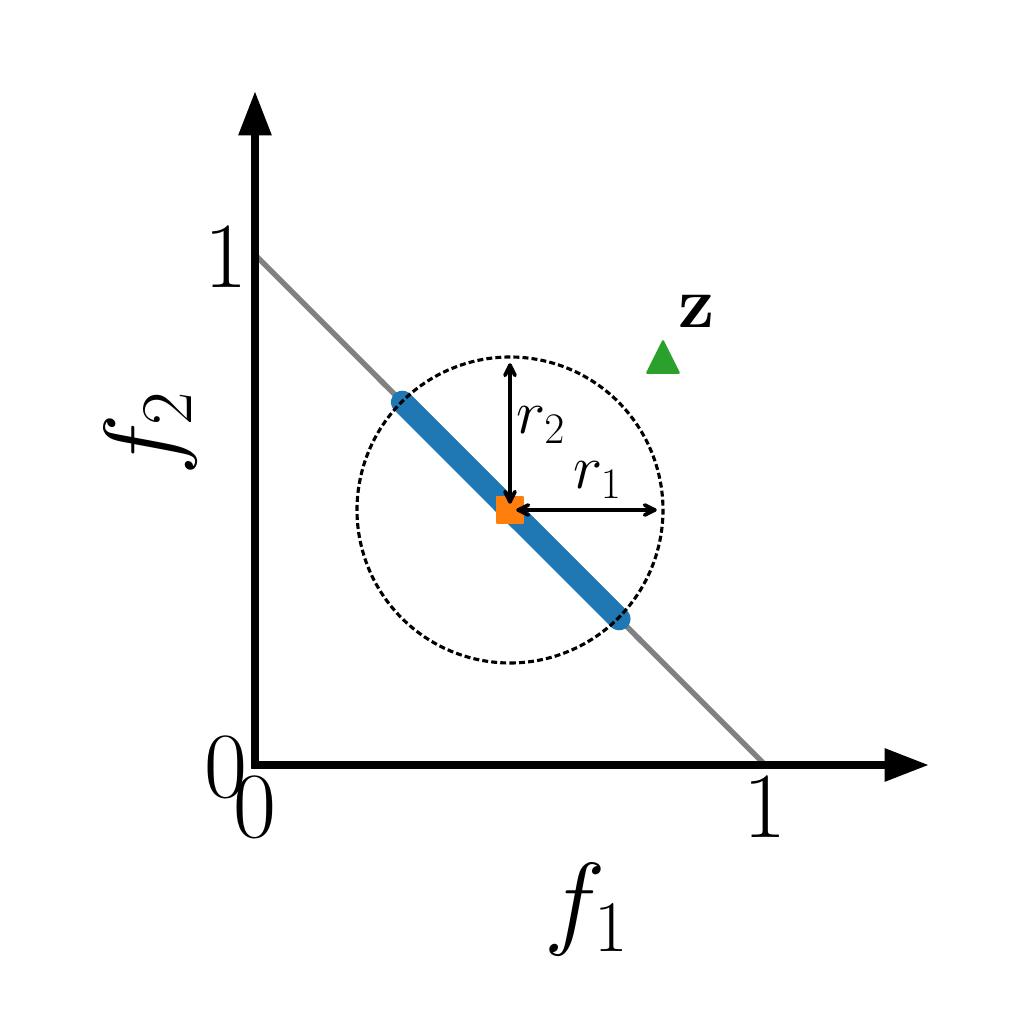}}
\subfloat[ROI-P]{\includegraphics[width=\wir\textwidth]{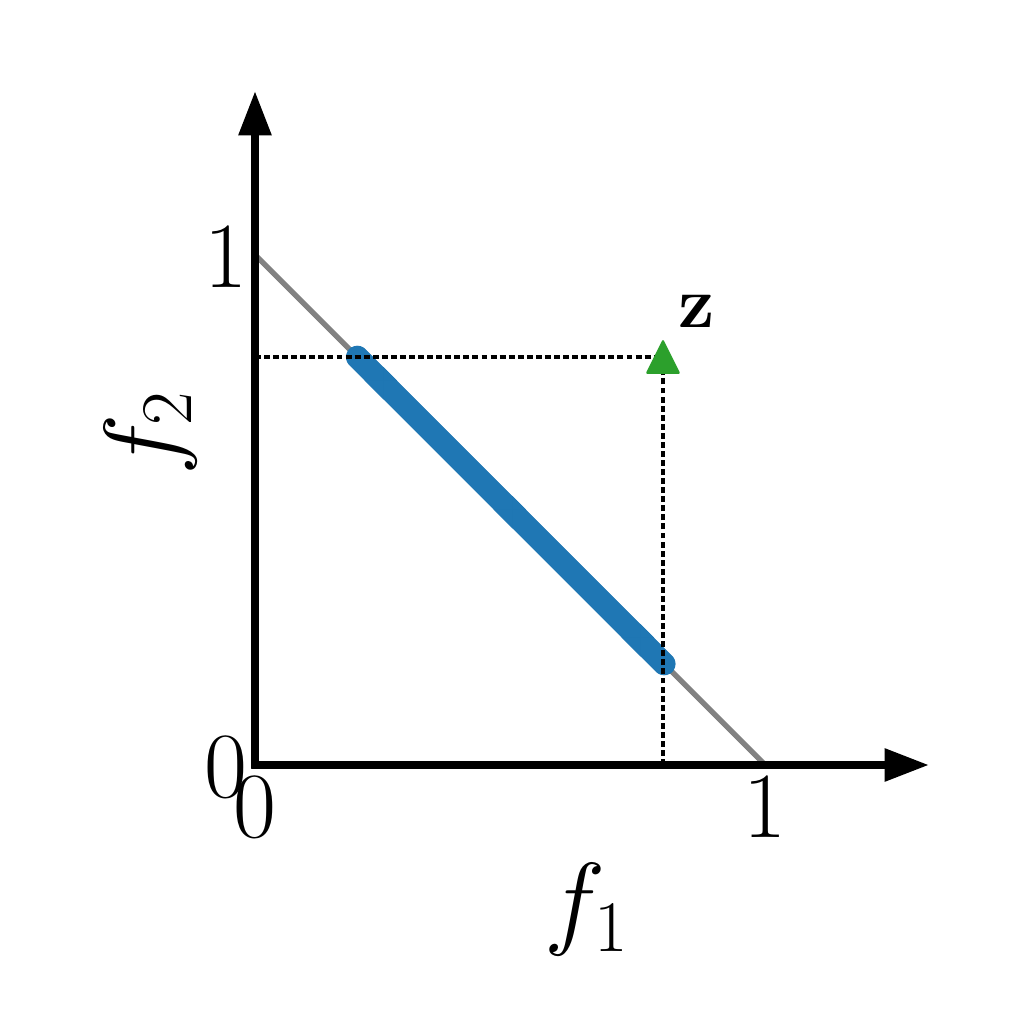}}
\caption{Examples of the ROI-C, ROI-A, and ROI-P, where {\color{nearpt}{\rule{5pt}{5pt}}} represents $\vec{f}(\vec{x}^{\text{c}*})$ or $\vec{f}(\vec{x}^{\text{a}*})$, and {\color{refpt}$\blacktriangle$} represents $\vec{z}$.
}
\label{fig:roi_example}
\end{figure}


\subsection{ROI}
\label{subsec:roi}


This section briefly describes the three ROIs (ROI-C, ROI-A, and ROI-P) defined in~\cite{TanabeL24}.
Each of the three ROIs is a subset of the PF defined by the reference point $\vec{z} = (z_1, \dots, z_m)^{\top}$.
The sizes of the ROI-C and ROI-A are specified by the semi-axis lengths $\vec{r}=(r_1, \dots, r_m)^{\top}$.\footnote{This is a generalized definition of the one used in the previous study~\cite{TanabeL24}, which assumes that all $m$ semi-axis lengths are identical.}
Both $\vec{z}$ and $\vec{r}$ are specified by the DM.
\pref{fig:roi_example} illustrates examples of the three ROIs, where $\vec{z}=(0.8, 0.8)^{\top}$ and $\vec{r}=(0.3, 0.3)^{\top}$.


\subsubsection*{ROI-C}
The ROI-C is defined as follows:
%
\begin{equation}
\label{eq:roi-c}
\begin{aligned}
\text{ROI-C}
&= \left\{
\vec f(\vec x^*) \in \vec f(\mathcal X^*)
\mid
\sum_{i=1}^{m}
\frac{\bigl(f_i(\vec{x}^*) - f_i(\vec{x}^{\mathrm{c}*})\bigr)^2}{r_i^2}
< 1
\right\},\\
\vec{x}^{\mathrm{c}*}
&= \underset{\vec{x}^* \in \mathcal{X}^*}{\operatorname{argmin}}
\left\{
\texttt{dist}\left(\vec{f}(\vec{x}^*), \vec{z}\right)
\right\}.
\end{aligned}
\end{equation}
where the function $\texttt{dist}$ returns the Euclidean distance between its two inputs.
In \pref{eq:roi-c}, $\vec{x}^{\text{c}*}$ is the Pareto-optimal solution closest to the reference point $\vec{z}$ in the objective space.


As shown in \pref{fig:roi_example} (a), the ROI-C is the set of Pareto-optimal vectors contained in an $m$-dimensional ellipsoid centered at $\vec{f}(\vec{x}^{\text{c}*})$ with semi-axis lengths $r_1, \dots, r_m$ in the objective space.
In principle, the $m$ elements of $\vec{r}$ can be set to different values; for example $\vec{r}=(0.1, 0.5)^{\top}$.
However, most previous studies (e.g.,~\cite{MohammadiOLD14,LiWTJE18,TanabeL24,MogamiT26}) assigned the same value to $r_1, \dots, r_m$.
In this special case, where all elements of $\vec{r}$ are equal to $r$, the ROI-C is the set of Pareto-optimal vectors contained in an $m$-dimensional hypersphere centered at $\vec{f}(\vec{x}^{\text{c}*})$ with radius $r$.


\subsubsection*{ROI-A}
The achievement scalarizing function (ASF)~\cite{Wierzbicki80} maps $m$ objective values $f_1 (\vec{x}), \dots, f_m (\vec{x})$ to a scalar value as follows:
\begin{align}
\label{eq:asf}
s(\vec{x}) = \max_{i \in \{1, ..., m\}} \bigl\{ w_i (f_i (\vec{x}) - z_i)  \bigr\},
\end{align}
where $s$ is the ASF, and $\mathbf{w}$ is the $m$-dimensional weight vector that indicates the relative importance of each objective function.
Here, $\sum^m_{i=1} w_i = 1$ and $w_i \geq 0$ for any $i$.
When the DM does not provide the relative importance, $\vec{w}$ is set to $(1/m, \dots, 1/m)^{\top}$.

The ROI-A is defined using the ASF function in \pref{eq:asf} as follows:
%
\begin{equation}
\label{eq:roi-a}
\begin{aligned}
\text{ROI-A}
&= \left\{
\vec f(\vec x^*) \in \vec f(\mathcal X^*)
\mid
\sum_{i=1}^{m}
\frac{\bigl(f_i(\vec{x}^*) - f_i(\vec{x}^{\mathrm{a}*})\bigr)^2}{r_i^2}
< 1
\right\},\\
\vec{x}^{\mathrm{a}*}
&= \underset{\vec{x}^* \in \mathcal{X}^*}{\operatorname{argmin}}
\left\{
s(\vec{x}^*)
\right\}.
\end{aligned}
\end{equation}
where $\vec{x}^{\text{a}*}$ is the Pareto-optimal solution with the minimum ASF value.
As shown in equations \eqref{eq:roi-c} and \eqref{eq:roi-a}, the difference between the ROI-C and ROI-A lies solely in the type of center vector.
However, except for the special case shown in \cite{TanabeL24}, $\vec{x}^{\text{c}*}$ and $\vec{x}^{\text{a}*}$ are the same.
In this case, the ROI-C and ROI-A are identical.


\subsubsection*{ROI-P}

The ROI-P is defined using the Pareto-dominance relation.
Below, the reference vector $\vec{z}$ is said to be feasible if $\vec{z}$ is dominated by at least one Pareto-optimal objective vector $\vec{f}(\vec{x}^*)$.
In contrast, $\vec{z}$ is said to be infeasible if $\vec{z}$ dominates at least one Pareto-optimal objective vector $\vec{f}(\vec{x}^*)$.
If $\vec{z}$ and all Pareto-optimal objective vectors are non-dominated, the ROI-P is an empty set.
This paper does not consider this case.

If $\vec{z}$ is feasible, the ROI-P is defined as the set of Pareto-optimal objective vectors that dominate $\vec{z}$, as follows:
\begin{equation}
\label{eq:nroip_f}
\text{ROI-P} = \{\vec{f}(\vec{x}^*) \in \vec{f}(\set{X}^*) \mid \vec{f}(\vec{x}^*) \prec \vec{z}\}. 
\end{equation}
%
the ROI-P is defined as the set of Pareto-optimal objective vectors that are dominated by $\vec{z}$, as follows:
\begin{equation}
\label{eq:nroip_if}
\text{ROI-P} = \{\vec{f}(\vec{x}^*) \in \vec{f}(\set{X}^*) \mid \vec{f}(\vec{x}^*) \succ \vec{z}\}. 
\end{equation}

\pref{fig:roi_example} (b) shows an example of the ROI-P when $\vec{z}$ is feasible.
A clear advantage of the ROI-P over the ROI-C and ROI-A is that it does not require specifying semi-axis lengths $r_1, \dots, r_m$, which may reduce the DM's burden.
However, in other words, the size of the ROI-P is uncontrollable.
For example, the ROI-P is identical to the PF when $\vec{z}$ dominates the ideal point or is dominated by the nadir point. 


\subsection{Normalization of objectives}
\label{subsec:normalization}

If the true ideal point $\vec{z}^\text{ideal}$ and nadir point $\vec{z}^\text{nadir}$ are known, the value of the $i$-th objective function $f_i$ can be normalized as follows:
%
\begin{align}
\label{eq:normalization}
    \hat{f}_i(\vec{x}) = \frac{f_i(\vec{x}) - z^\text{ideal}_i}{z^\text{nadir}_i - z^\text{ideal}_i}.
\end{align}
The normalized objective values $\hat{f}_1(\vec{x}), \ldots, \hat{f}_m(\vec{x})$ have the same scale. 
However, the true $\vec{z}^\text{ideal}$ and $\vec{z}^\text{nadir}$ are unknown in real-world applications.
Therefore, objective normalization in \pref{eq:normalization} cannot be used during the evolutionary search.
For this reason, EMO algorithms generally normalize the objective values by using approximations of $\vec{z}^\text{ideal}$ and $\vec{z}^\text{nadir}$ as follows:
\begin{align}
\label{eq:normalization_2}
    \hat{f}_i(\vec{x}) = \frac{f_i(\vec{x}) - z^\text{lb}_i}{z^\text{ub}_i - z^\text{lb}_i}, 
\end{align}
where $\vec{z}^\text{lb} \in \mathbb{R}^m$ and $\vec{z} ^\text{ub} \in \mathbb{R}^m$ are approximations of $\vec{z}^\text{ideal}$ and $\vec{z}^\text{nadir}$, respectively.
As reviewed in~\cite{HeITWNS21}, various methods have been proposed to estimate $\vec{z}^\text{lb}$ and $\vec{z}^\text{ub}$.
For example, at each iteration $t$, the simplest method estimates $\vec{z}^\text{lb}$ and $\vec{z}^\text{ub}$ from the current population $\set{P}$ as follows: $z^\text{lb}_i = \min_{\vec{x} \in \set{P}} f_i(\vec{x})$, and $z^\text{ub}_i = \max_{\vec{x} \in \set{P}} f_i(\vec{x})$.


%% file: roi_definition.tex

\section{Effects of objective normalization on ROIs}
\label{sec:roi_definition}


This section examines the differences between ROIs defined in the unnormalized objective space $\mathbb{R}^m$ and the normalized objective space $[0,1]^m$ described in \pref{sec:introduction}.
For simplicity, the former is referred to as the unnormalized ROI (UROI), whereas the latter is referred to as the normalized ROI (NROI).
The ROI-C, ROI-A, and ROI-P described in \pref{subsec:roi} are all classified as UROIs.
In the following, they are referred to as the UROI-C, UROI-A, and UROI-P, respectively.
Their normalized versions (the NROI-C, NROI-A, and NROI-P) can be obtained by simply replacing $\vec{f}(\vec{x}^*)$ with $\hat{\vec{f}}(\vec{x}^*)$ and $\vec{z}$ with $\hat{\vec{z}}$.
Here, $\hat{\vec{f}}(\vec{x}^*)$ and $\hat{\vec{z}}$ are the normalized Pareto-optimal objective vector and the normalized reference vector obtained using \pref{eq:normalization}, respectively.
Thus, normalization of objectives is performed using the true ideal and nadir points. 
In summary, this section addresses the three UROIs (the UROI-C, UROI-A, and UROI-P) and the three NROIs (the NROI-C, NROI-A, and NROI-P).
Note that, as shown in \pref{tab:type_roi}, no consensus exists in the EMO community as to which of the six ROI types should be used.
Note also that this section focuses on the effects of objective normalization on ROIs, not on the behavior of PBEMO algorithms.


\begin{figure}[t]
\newcommand{\wir}{0.32}
\centering
\subfloat[UROI-C and UROI-A]{\includegraphics[width=\wir\textwidth]{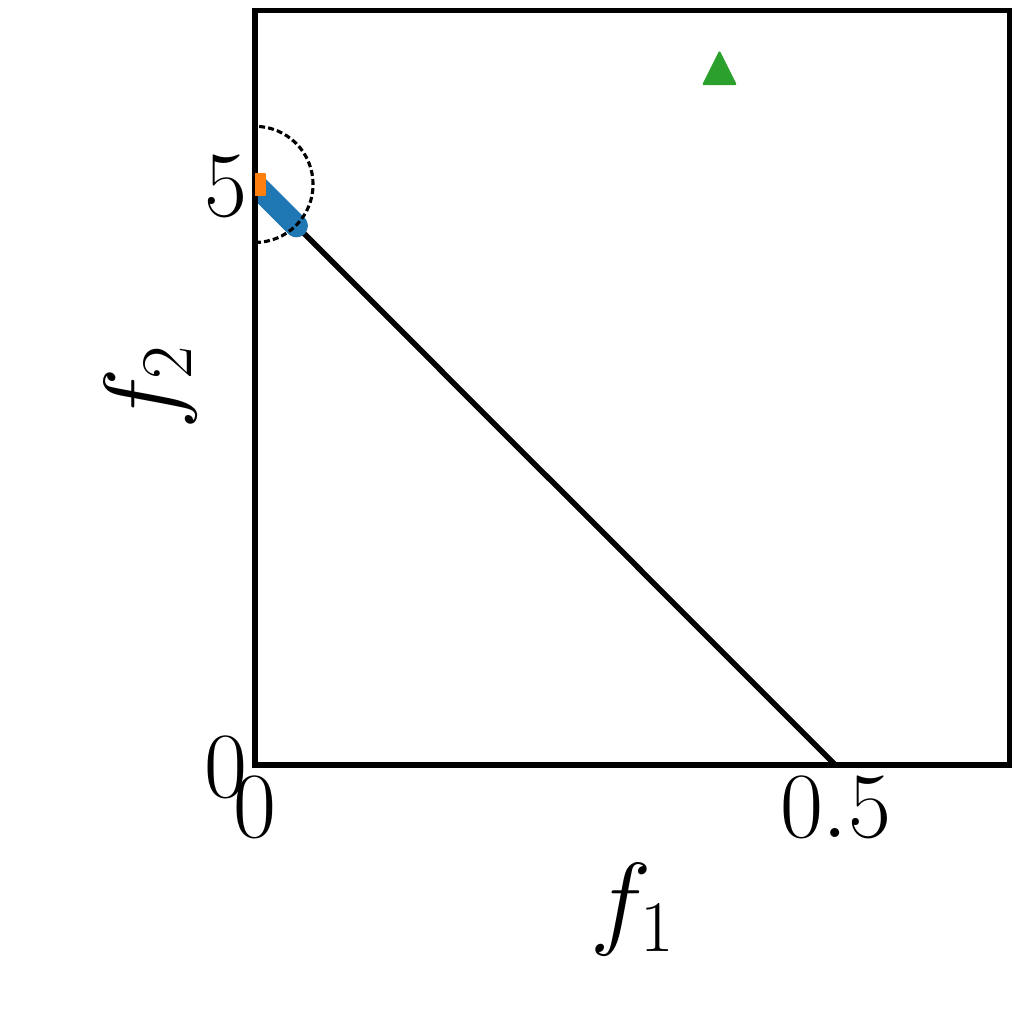}}
\subfloat[NROI-C and NROI-A]{\includegraphics[width=\wir\textwidth]{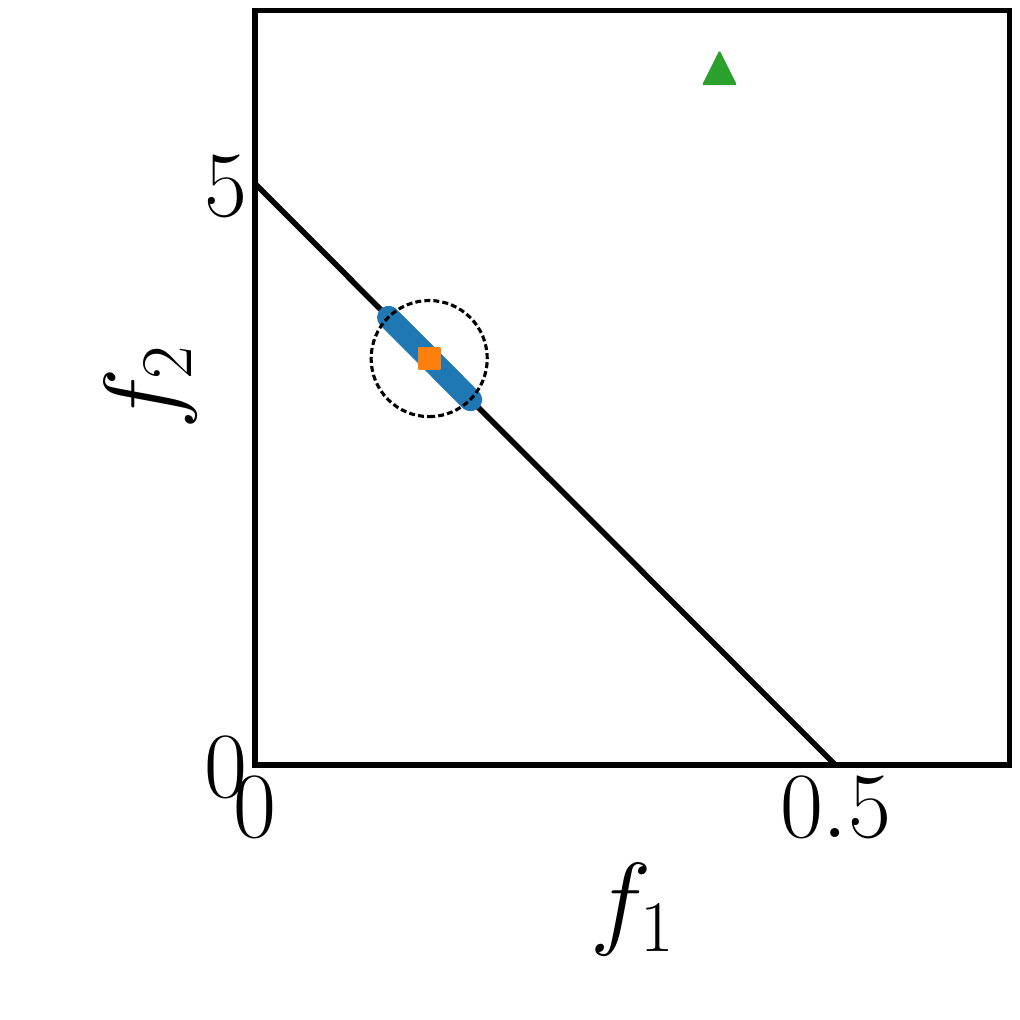}}
\subfloat[UROI-P and NROI-P]{\includegraphics[width=\wir\textwidth]{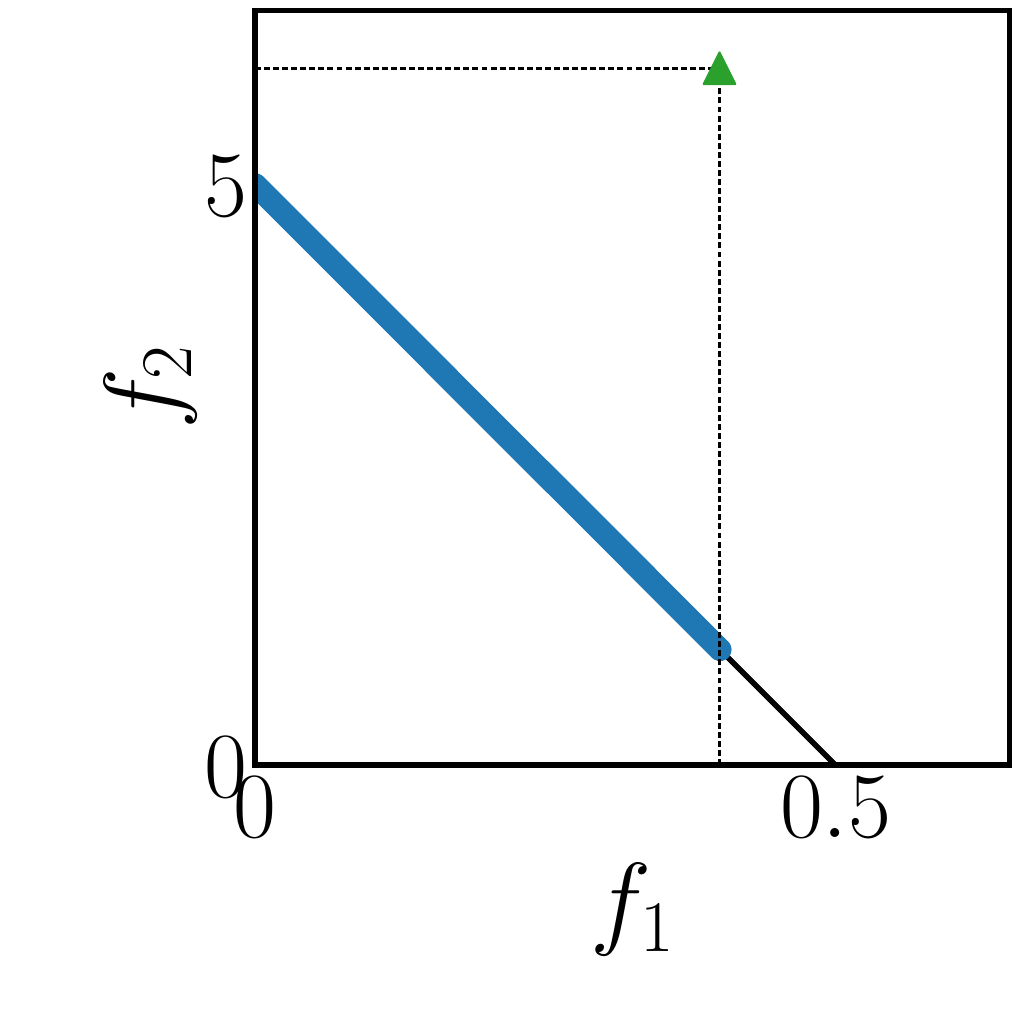}}
\caption{Distributions of the six ROIs on the SDTLZ1 problem. 
}
\label{fig:roica_demo_sdtlz}
\end{figure}

\begin{figure}[t]
\newcommand{\wir}{0.3}
\centering
\subfloat[Unnormalized space]{\includegraphics[width=\wir\textwidth]{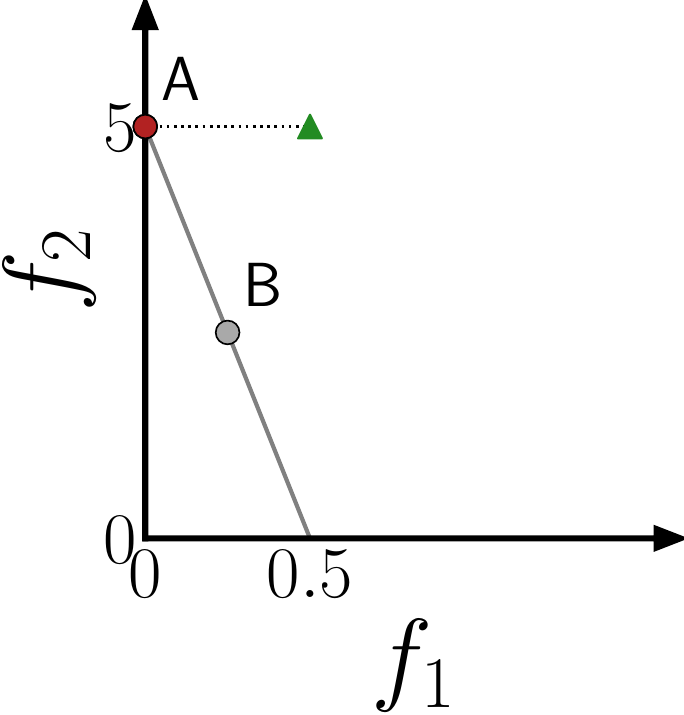}}
\subfloat[Normalized space]{\includegraphics[width=\wir\textwidth]{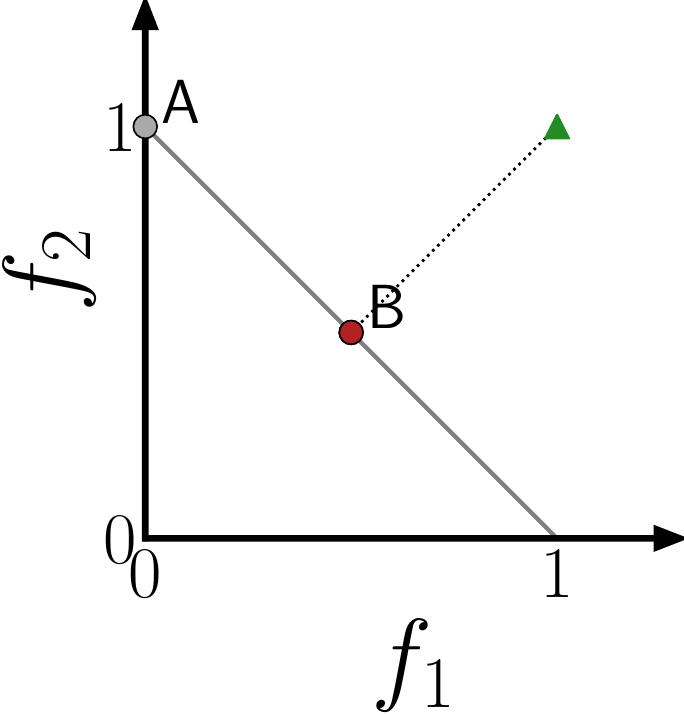}}
\caption{Distributions of the PFs of the bi-objective SDTLZ1 problem for the unnormalized and normalized objective spaces.
}
\label{fig:roic_euc}
\end{figure}





\pref{fig:roica_demo_sdtlz} shows distributions of the six ROIs for the bi-objective SDTLZ1 problem, in which the scale of $f_2$ is $10$ times larger than that of $f_1$.
For the bi-objective SDTLZ1 problem, the ideal point $\vec{z}^\text{ideal}$ and nadir point $\vec{z}^\text{nadir}$ are as follows: $\vec{z}^\text{ideal}=(0, 0)^\top$ and $\vec{z}^\text{nadir}=(0.5, 5)^\top$.
In the example of \pref{fig:roica_demo_sdtlz}, the reference vector $\vec{z}$ is given by $\vec{z} = (0.4, 6)^\top$.
Semi-axis lengths $\vec{r}$ are $\vec{r} = (0.05, 0.5)^{\top}$ for the three UROIs and $\hat{\vec{r}} = (0.1, 0.1)^\top$ for the NROIs.
We used these two settings of $\vec{r}$ so that the UROI-C, UROI-A, NROI-C, and NROI-A had the same size in the original, unnormalized objective space.
In this example, the distributions of the UROI-C and UROI-A in \pref{fig:roica_demo_sdtlz}(a) are exactly the same.
The same holds for the NROI-C and NROI-A in \pref{fig:roica_demo_sdtlz}(b), and for the UROI-P and NROI-P in \pref{fig:roica_demo_sdtlz}(c).
Figs. \ref{fig:roica_demo_sdtlz}(b) and (c) show the mappings of the three NROIs to the original unnormalized objective space.

As shown in \pref{fig:roica_demo_sdtlz}(a) and \pref{fig:roica_demo_sdtlz}(b), the positions of the UROI-C and NROI-C are totally different.
The same is true for the UROI-A and NROI-A.
While the UROI-C and UROI-A are distributed around $(0.0, 5.0)^\top$, the NROI-C and NROI-A are distributed around $(0.15, 3.5)^\top$.
Similar results are also observed for other settings of $\vec{z}$ and for other problems with differently scaled objectives, including the SDTLZ2--4 and WFG1--9 problems, when $m \in \{2, 3\}$.
In contrast, as seen from \pref{fig:roica_demo_sdtlz}(c), the distributions of the UROI-P and NROI-P are the same.
Since the ROI-P is defined based on the Pareto-dominance relation, it relies only on the relative ordering of the objective values.
For this reason, the ROI-P is not influenced by whether the objectives are normalized.


As described above, we observed that the distributions of the UROI-C and NROI-C are different, and the same holds for those of the UROI-A and NROI-A.
This is because the center points $\vec{x}^{\mathrm{c}*}$ and $\vec{x}^{\mathrm{a}*}$ in equations \eqref{eq:roi-c} and \eqref{eq:roi-a} can differ depending on whether the ROI is defined in the normalized or unnormalized objective space.

Figs. \ref{fig:roic_euc} (a) and (b) show the PFs of the bi-objective SDTLZ1 problem for the unnormalized and normalized objective spaces, respectively.
In \pref{fig:roic_euc}, the symbol {\color{refpt}$\blacktriangle$} represents the reference point $\vec{z}$, and the two symbols {\color{candpt}$\bullet$} represent the two candidates for $\vec{x}^{\mathrm{c}*}$ and $\vec{x}^{\mathrm{a}*}$. 
As shown in \pref{fig:roic_euc} (a), Candidate A is selected as $\vec{x}^{\mathrm{c}*}$ and $\vec{x}^{\mathrm{a}*}$ in the unnormalized objective space.
Candidate A clearly minimizes both the distance to $\vec{z}$ and the ASF value in \pref{eq:asf}.
In contrast, as seen from \pref{fig:roic_euc} (b), Candidate B is selected as $\vec{x}^{\mathrm{c}*}$ and $\vec{x}^{\mathrm{a}*}$ in the normalized objective space $[0,1]^2$.
In \pref{fig:roic_euc}(b), Candidate A is the farthest from $\vec{z}$ and has the worst ASF value.
As demonstrated here, when the objective functions have different scales, the relative rankings of the candidates with respect to the distance to $\vec{z}$ and the ASF value significantly depend on whether the objective space is normalized or unnormalized.
This leads to the different distributions shown in Figs. \ref{fig:roica_demo_sdtlz}(a) and (b).

This is problematic for real-world applications of PBEMO.
If the DM is unaware of the effects of objective normalization on the ROIs, they are likely to obtain a solution set that is far from their expectation in the objective space.
For example, suppose that the DM implicitly seeks a good approximation of the NROI-A and uses MOEA/D-NUMS~\cite{LiCMY18} for this purpose.
In this case, the DM is unlikely to obtain the desired solution set, because MOEA/D-NUMS aims to approximate the UROI-A, as shown in \pref{tab:type_roi}.
The DM may incorrectly conclude that MOEA/D-NUMS performs poorly on the real-world problem.


In addition, our findings have a serious impact on the benchmarking of PBEMO algorithms.
Let us consider the benchmarking of R-NSGA-II~\cite{DebS06} and R-MEAD2~\cite{MohammadiOLD14}.
As shown in \pref{tab:type_roi}, while R-NSGA-II aims to approximate the NROI-C, R-MEAD2 aims to approximate the UROI-C.
If practitioners performing the benchmarking consider the NROI-C, R-NSGA-II is likely to outperform R-MEAD2.
In contrast, if they consider the UROI-C, R-NSGA-II is likely to be outperformed by R-MEAD2.
This is simply because R-NSGA-II and R-MEAD2 approximate different ROIs. Although this is a fundamental issue in benchmarking, it has been overlooked for a long time.
As shown in \pref{tab:type_roi}, there is no standard as to which objective space the ROI should be defined in.

\begin{tcolorbox}[sharpish corners, top=2pt, bottom=2pt, left=4pt, right=4pt, boxrule=0.0pt, colback=black!5!white,leftrule=0.75mm,]
\textbf{Summary and discussion}:
We demonstrated that the distribution of the ROI-C can differ significantly for problems with differently scaled objectives, depending on whether it is defined in the unnormalized or normalized objective space.
Although the same is true for the ROI-A, the ROI-P is not affected by the choice of space.
In light of this, the DM and practitioners conducting the benchmarking should define the ROI to be approximated before selecting and running PBEMO algorithms.
Otherwise, they are likely to obtain neither a desirable solution set nor reliable benchmarking results.

\end{tcolorbox}


%% file: setup.tex
\section{Experimental setup}
\label{sec:setup}

This section describes the experimental setup for our analysis.
We consider the NROI-C and UROI-C.
Since the ROI-C and ROI-A are equivalent in our experimental setting, we omit the ROI-A.

We used R-NSGA-II~\cite{DebS06}, which is one of the most representative PBEMO algorithms.
As shown in \pref{tab:type_roi}, R-NSGA-II was designed to approximate the NROI-C.
R-NSGA-II normalizes the objective values using the minimum and maximum objective values in each nondominated front.
For the analysis, we consider R-NSGA-II without this normalization mechanism.
For the sake of readability, we refer to the original and unnormalized versions of R-NSGA-II as NR-NSGA-II and UR-NSGA-II, respectively.
The \texttt{pymoo}~\cite{BlankD20} implementation of R-NSGA-II was used in this work.
In our preliminary experiments, we investigated the behavior of other PBEMO algorithms for the ROI-C and ROI-A, including r-NSGA-II~\cite{SaidBG10}, MOEA/D-NUMS~\cite{LiCMY18}, PBEA~\cite{ThieleMKL09}, R-MEAD2~\cite{MohammadiOLD14}, B-NSGA-II~\cite{MogamiT26}, B-IBEA~\cite{MogamiT26}, and B-SMS-EMOA~\cite{MogamiT26}.
Although their results are omitted due to space limitations, their behavior was similar to that of R-NSGA-II.
We performed 31 independent runs for each problem. 
The maximum number of function evaluations was set to $50\,000$.
The population size $\mu$ was set to 100. 
Parameters for UR-NSGA-II and NR-NSGA-II were set to default values~\cite{BlankD20}.


%

The DTLZ1--DTLZ4~\cite{DebTLZ05} and SDTLZ1--SDTLZ4~\cite{DebJ14} problems were used.
Here, SDTLZ$*$ is a modified version of DTLZ$*$ in which the objective functions are differently scaled by multiplying each objective function $f_i$ by $10^{i-1}$ for $i \in \{1, \ldots, m\}$.
For ease of visualization, the number of objectives, $m$, was set to $2$.
The number of decision variables was set as in~\cite{DebTLZ05}.
The reference point $\vec{z}$ was set as in~\cite{MogamiT26}.
The semi-axis lengths $\vec{r}$ were set such that $\vec{r} = (0.1, \dots, 0.1)^{\top}$ in the normalized objective space.
For example, $\vec{r} = (0.1, 1)^{\top}$ for the SDTLZ2 problem with $m=2$.

$\text{IGD}^+\text{-C}$~\cite{Tanabe23} was used to evaluate the quality of solution sets for the ROI-C.
$\text{IGD}^+$~\cite{IshibuchiMTN15} is a weakly Pareto-compliant version of inverted generational distance (IGD)~\cite{CoelloS04}, which evaluates the quality of solution sets in terms of convergence and diversity with respect to the PF.
$\text{IGD}^+\text{-C}$ is an extended version of $\text{IGD}^+$ to evaluate how well solution sets approximate the ROI-C.
The difference between IGD$^+$ and $\text{IGD}^+\text{-C}$ is only the distribution of the IGD-reference point set $\set{S}$ in the objective space.
Unlike points in $\set{S}$ for IGD$^+$, those for $\text{IGD}^+\text{-C}$ are distributed only on the ROI-C.
A small $\text{IGD}^+\text{-C}$ value indicates that the solution set approximates the ROI-C well in the objective space in terms of both convergence and diversity.


%% file: results.tex
\newcommand{\colortextbox}[2]{%
  \begingroup
  \setlength{\fboxsep}{1pt}%
  \colorbox{#1}{\strut #2}%
  \endgroup
}

\section{Results}
\label{sec:results}

\begin{table}[t]
  \centering
  \footnotesize
  \caption{Pair-wise comparison between UR-NSGA-II and NR-NSGA-II on the bi-objective DTLZ and SDTLZ problems.}
  \label{tab:roic_comparison}

  \subfloat[UROI-C\label{tab:roic_comparison_a}]{
    \resizebox{.4\linewidth}{!}{%
    \begin{tabular}{c c c}
      \toprule
      Problem & UR-NSGA-II & NR-NSGA-II \\
      \midrule
      DTLZ1  & \cellcolor{black!20}0.0052 & 0.0189 \\
      DTLZ2  & \cellcolor{black!20}0.0085 & 0.0919 \\
      DTLZ3  & \cellcolor{black!20}0.0127 & 0.0905 \\
      DTLZ4  & 0.0598 & \cellcolor{black!20}0.0454 \\
      SDTLZ1 & \cellcolor{black!20}0.0016 & 0.2569 \\
      SDTLZ2 & \cellcolor{black!20}0.0034 & 0.2422 \\
      SDTLZ3 & \cellcolor{black!20}0.0048 & 0.2549 \\
      SDTLZ4 & \cellcolor{black!20}0.1222 & 0.4078 \\
      \bottomrule
    \end{tabular}%
    }
  }
  \hfill
  \subfloat[NROI-C\label{tab:roic_comparison_b}]{
    \resizebox{.4\linewidth}{!}{%
    \begin{tabular}{c c c}
      \toprule
      Problem & UR-NSGA-II & NR-NSGA-II \\
      \midrule
      DTLZ1  & \cellcolor{black!20}0.0052 & 0.0189 \\
      DTLZ2  & \cellcolor{black!20}0.0085 & 0.0919 \\
      DTLZ3  & \cellcolor{black!20}0.0127 & 0.0905 \\
      DTLZ4  & 0.0598 & \cellcolor{black!20}0.0454 \\
      SDTLZ1 & 0.2635 & \cellcolor{black!20}0.0189 \\
      SDTLZ2 & 0.1856 & \cellcolor{black!20}0.0964 \\
      SDTLZ3 & 0.1831 & \cellcolor{black!20}0.1005 \\
      SDTLZ4 & 0.2094 & \cellcolor{black!20}0.0571 \\
      \bottomrule
    \end{tabular}%
    }
  }
\end{table}


Whereas \pref{sec:roi_definition} examined the effects of objective normalization on ROIs, this section investigates the performance of UR-NSGA-II and NR-NSGA-II in approximating the UROI-C and NROI-C.
\pref{tab:roic_comparison} shows the pair-wise comparison between UR-NSGA-II and NR-NSGA-II on the bi-objective DTLZ1--4 and SDTLZ1--4 problems for the UROI-C and NROI-C. \pref{tab:roic_comparison} shows the average $\text{IGD}^+\text{-C}$ values of the final populations of each PBEMO algorithm over 31 runs. 
Data highlighted in \colortextbox{lightgray}{gray} indicate that the corresponding algorithm performs significantly better than the other according to the Wilcoxon rank-sum test with $p < 0.05$. 
As demonstrated in \pref{sec:roi_definition}, the UROI-C and NROI-C are identical for problems with objectives on the same scale, such as the DTLZ1--4 problems.
For this reason, the results for the DTLZ1--4 problems are the same in Tables \ref{tab:roic_comparison}(a) and (b).

Since the UROI-C and NROI-C are identical for the DTLZ problems, as shown in \pref{sec:roi_definition}, UR-NSGA-II and NR-NSGA-II aim to approximate the same ROI.
Therefore, UR-NSGA-II and NR-NSGA-II are expected to perform similarly on the DTLZ problems.
However, as shown in \pref{tab:roic_comparison}, UR-NSGA-II achieves better $\text{IGD}^+\text{-C}$ values than NR-NSGA-II on the DTLZ1--3 problems. 
The reason for this counterintuitive result is examined in \pref{subsec:discussion_dtlz}.

As shown in \pref{tab:roic_comparison}, as expected, UR-NSGA-II outperforms NR-NSGA-II in approximating the UROI-C on the SDTLZ1--4 problems, whereas NR-NSGA-II outperforms UR-NSGA-II in approximating the NROI-C.
\pref{subsec:discussion_sdtlz} analyzes these results.


\subsection{Why does NR-NSGA-II perform poorly on the DTLZ problems?}
\label{subsec:discussion_dtlz}

\begin{figure}[t]
\newcommand{\wir}{0.32}
\centering
\subfloat[UR-NSGA-II]{\includegraphics[width=\wir\textwidth]{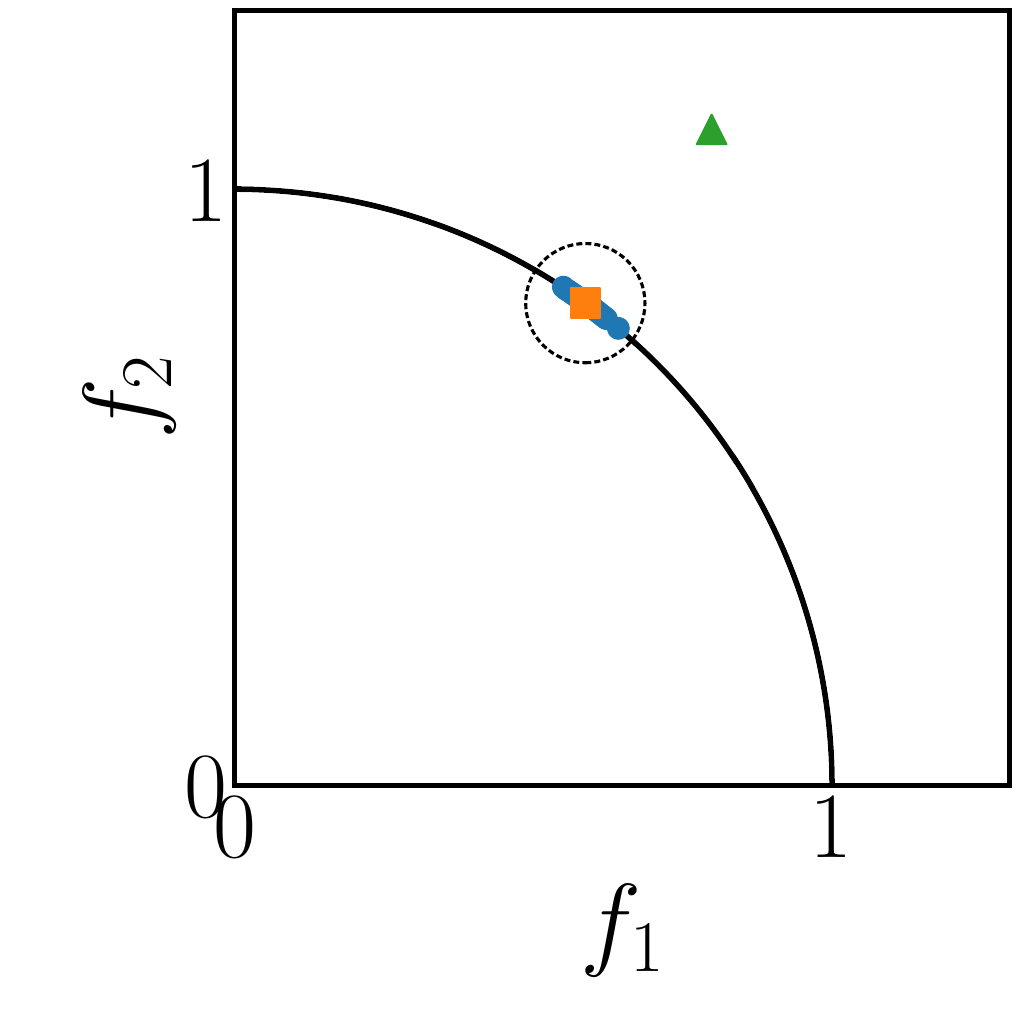}}
\subfloat[NR-NSGA-II]{\includegraphics[width=\wir\textwidth]{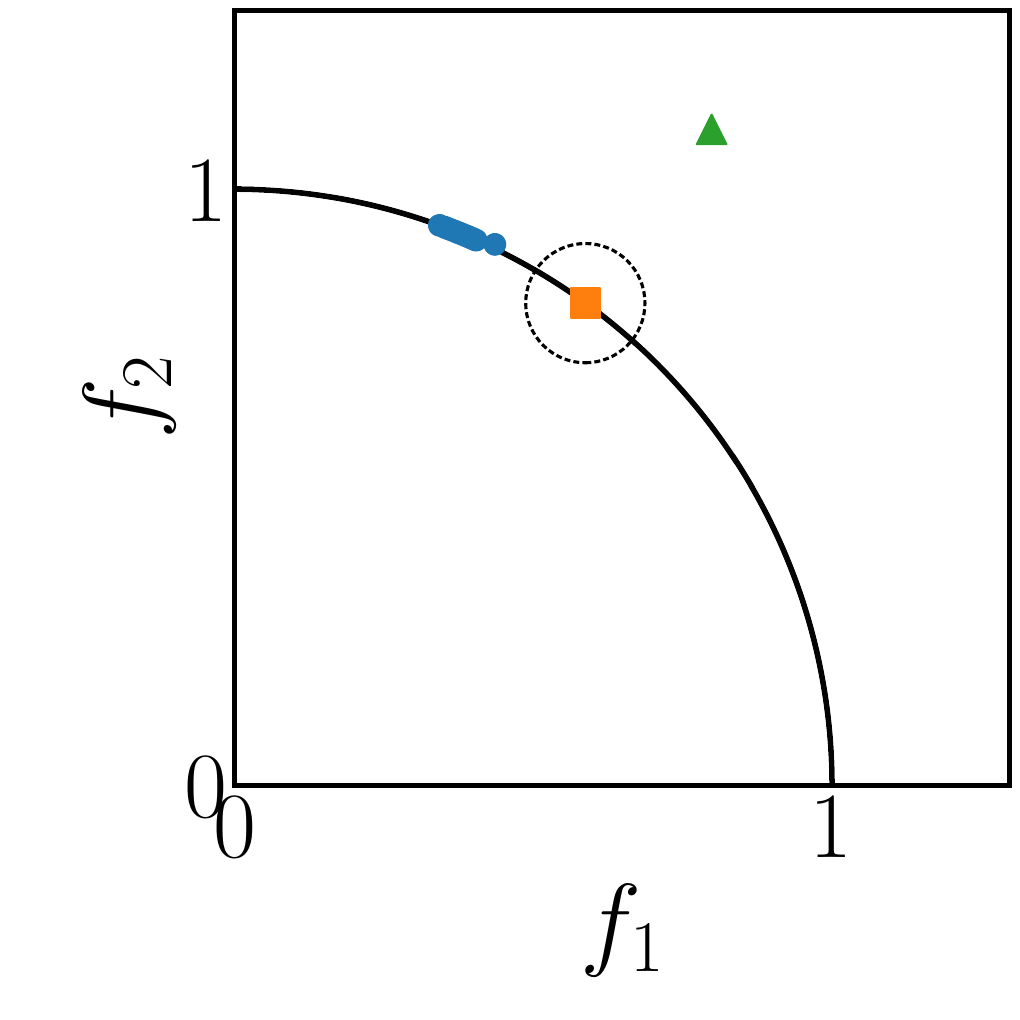}}
\caption{Distributions of the final populations of UR-NSGA-II and NR-NSGA-II on the DTLZ2 problem, where {\color{nearpt}{\rule{5pt}{5pt}}} is the center of the UROI-C and NROI-C, and {\color{refpt}$\blacktriangle$} represents the reference point $\vec{z}$.
}
\label{fig:roic_dtlz2}
\end{figure}

\pref{fig:roic_dtlz2} shows the final populations of UR-NSGA-II and NR-NSGA-II on the bi-objective DTLZ2 problem.
The results are from the median-IGD$^+$-C run among the 31 runs.
As shown in \pref{fig:roic_dtlz2}(a), the final population of UR-NSGA-II approximates the ROI-C well.
In contrast, the final population of NR-NSGA-II is distributed on the PF but remains far from the ROI-C.

\begin{figure}[t]
\newcommand{\wir}{0.32}
\centering
\subfloat[$1\,000$ \texttt{FE}]{\includegraphics[width=\wir\textwidth]{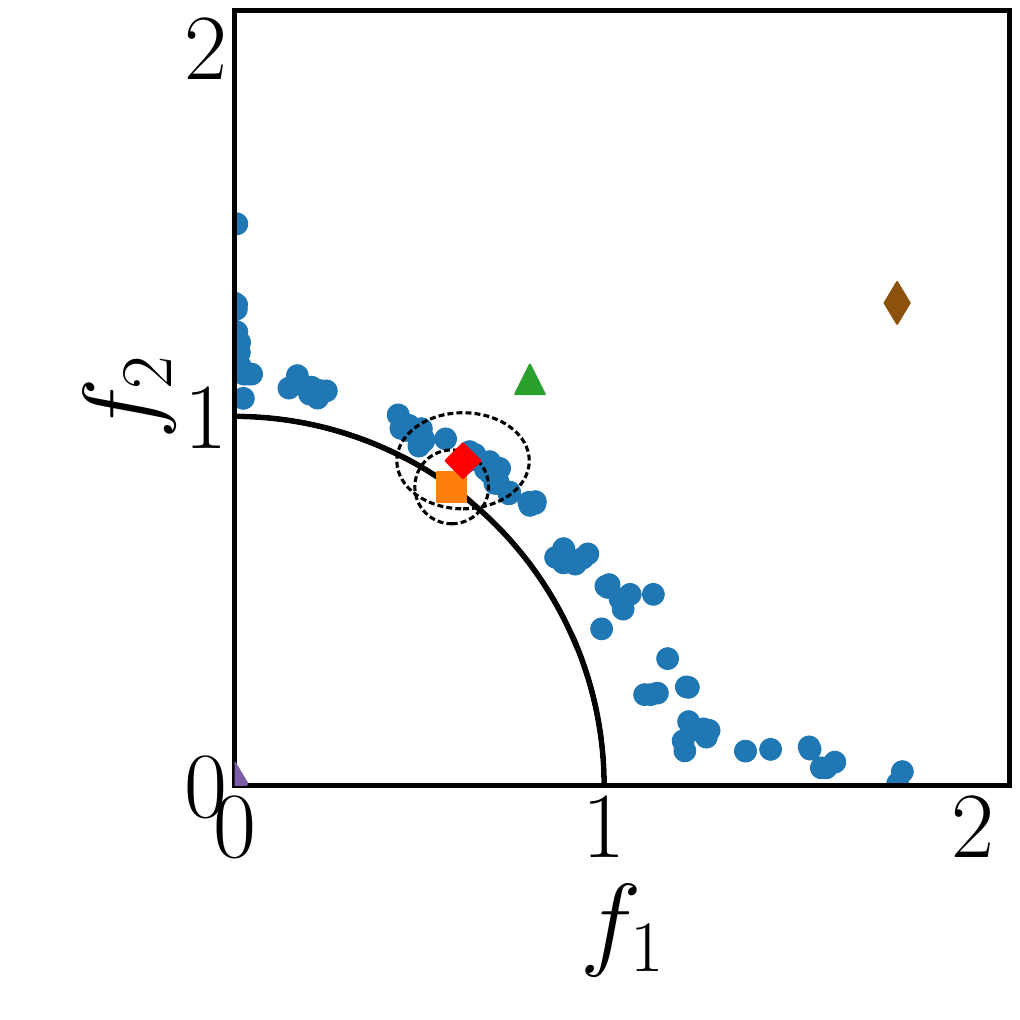}}
\subfloat[$3\,000$ \texttt{FE}]{\includegraphics[width=\wir\textwidth]{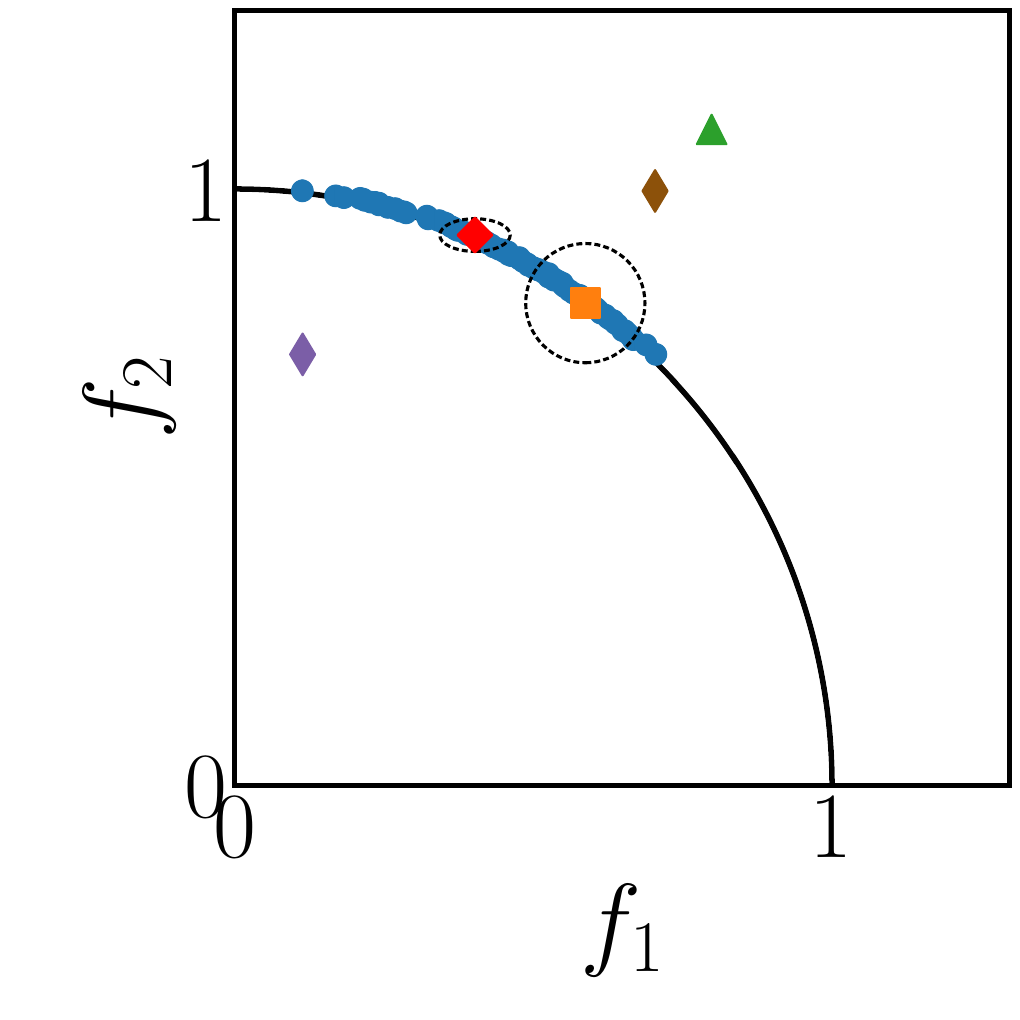}}
\subfloat[$5\,000$ \texttt{FE}]{\includegraphics[width=\wir\textwidth]{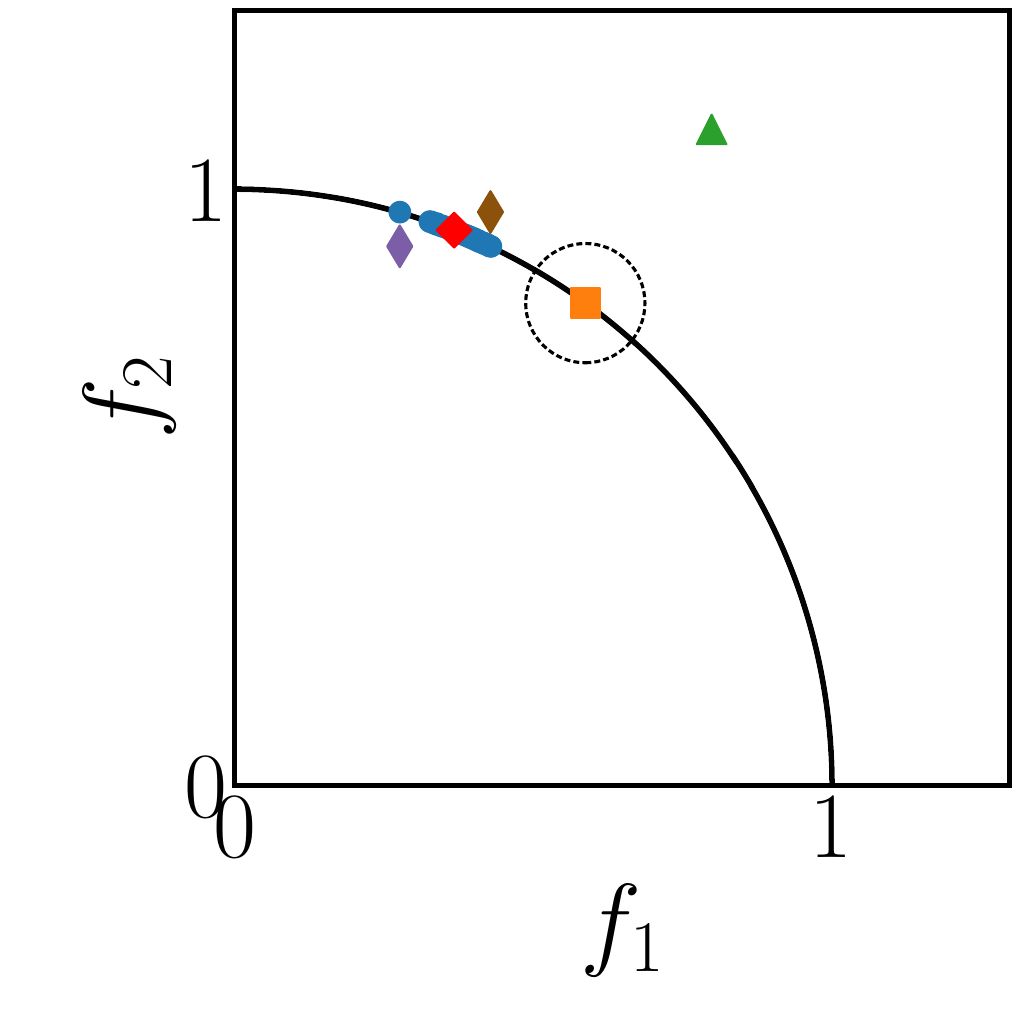}}\\
\caption{Distributions of the populations in NR-NSGA-II on the DTLZ2 problem in the \textit{unnormalized} objective space, where {\color{ideal}$\blacklozenge$} and {\color{nadir}$\blacklozenge$} represent $\vec{z}^\mathrm{lb}$ and $\vec{z}^\mathrm{ub}$, respectively.
}
\label{fig:nrnsga2_dtlz2_progress}
%
\vspace{1em}
\subfloat[$1\,000$ \texttt{FE}]{\includegraphics[width=\wir\textwidth]{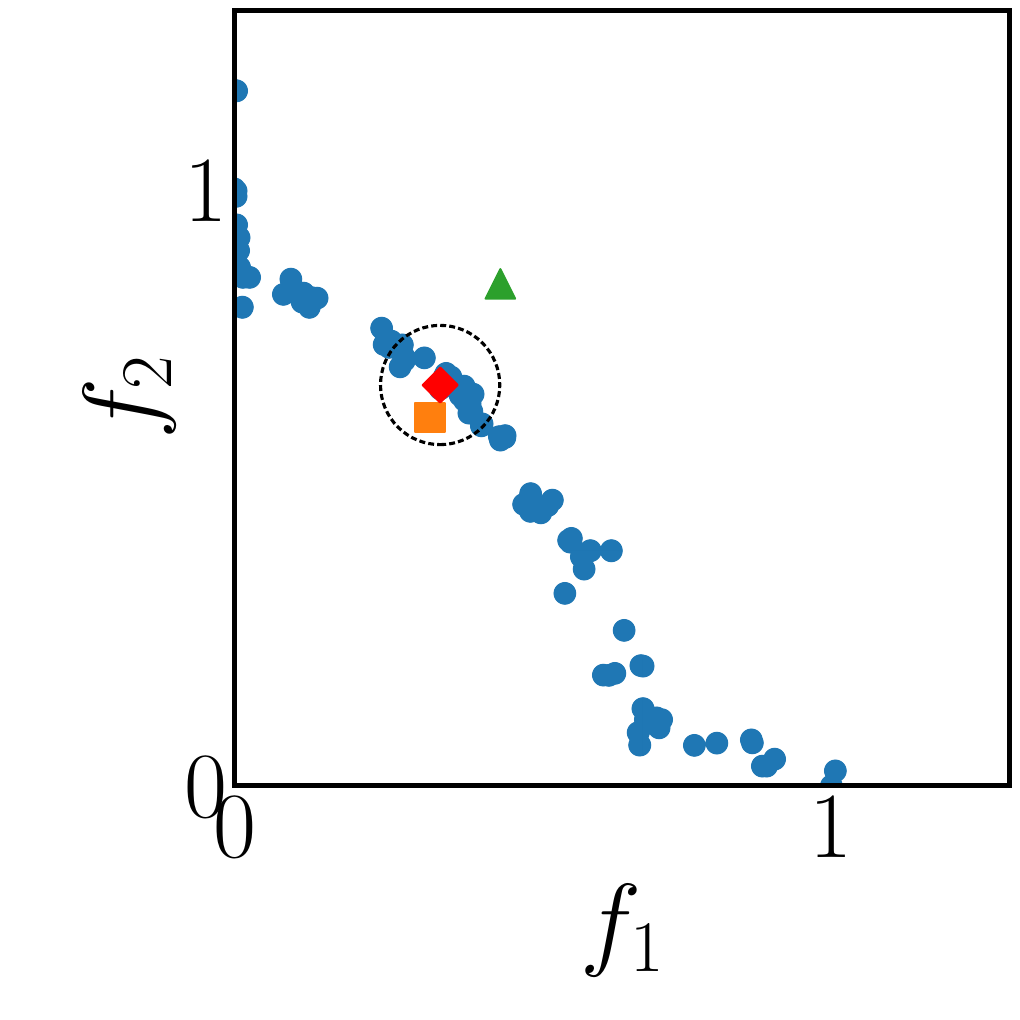}}
\subfloat[$3\,000$ \texttt{FE}]{\includegraphics[width=\wir\textwidth]{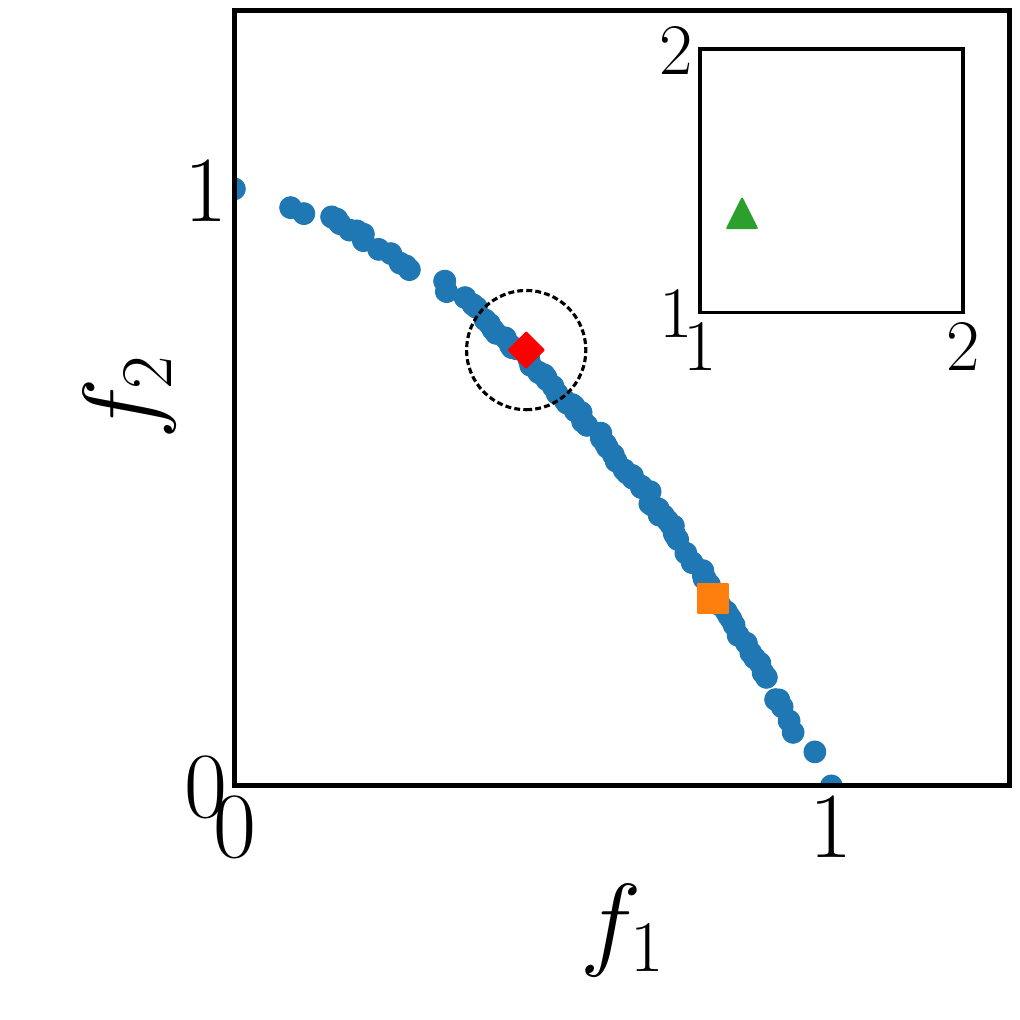}}
\subfloat[$5\,000$ \texttt{FE}]{\includegraphics[width=\wir\textwidth]{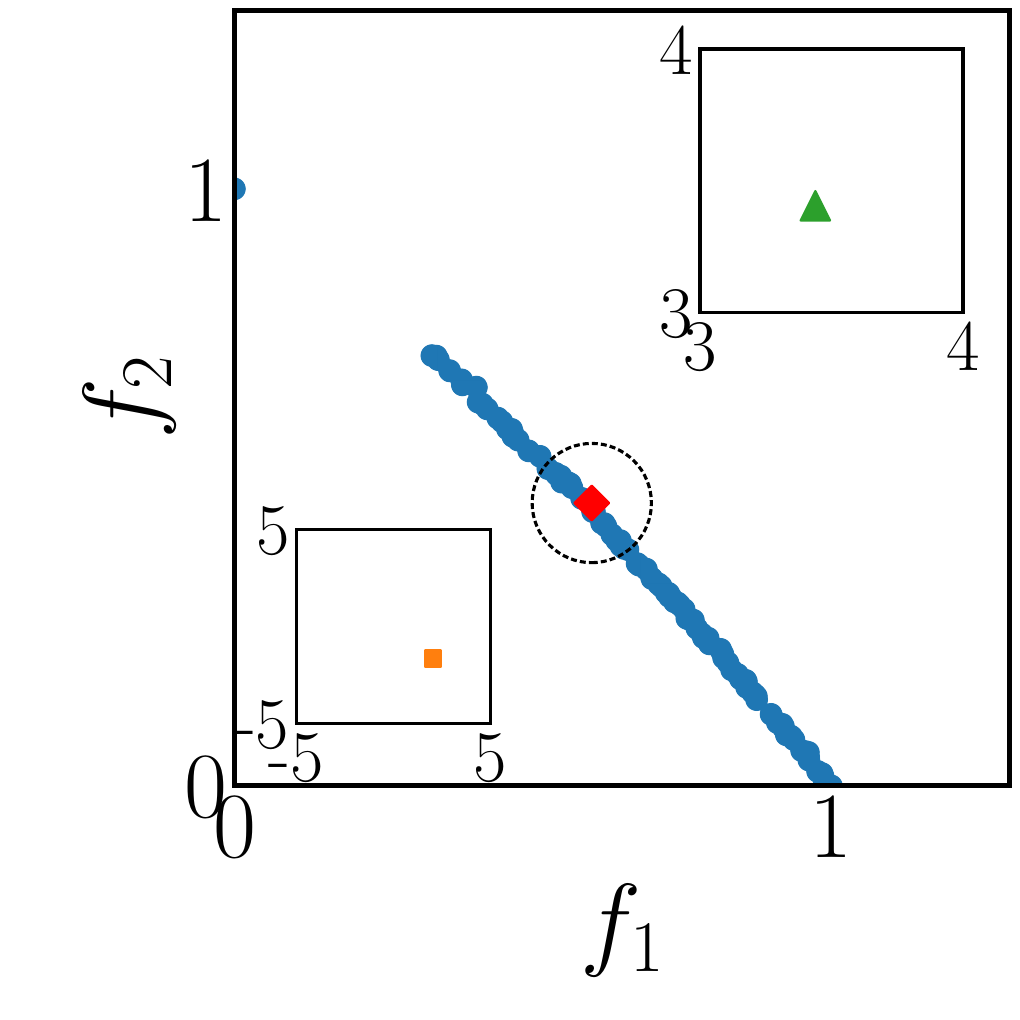}}\\
\caption{Distributions of the populations in NR-NSGA-II on the DTLZ2 problem in the \textit{normalized} objective space, where $\vec{z}^\mathrm{lb}$ and $\vec{z}^\mathrm{ub}$ are used for objective normalization.
}
\label{fig:nrnsga2_dtlz2_progress_normalized}
\end{figure}

\begin{figure}[t]
\newcommand{\wir}{0.32}
\centering
\subfloat[UR-NSGA-II]{\includegraphics[width=\wir\textwidth]{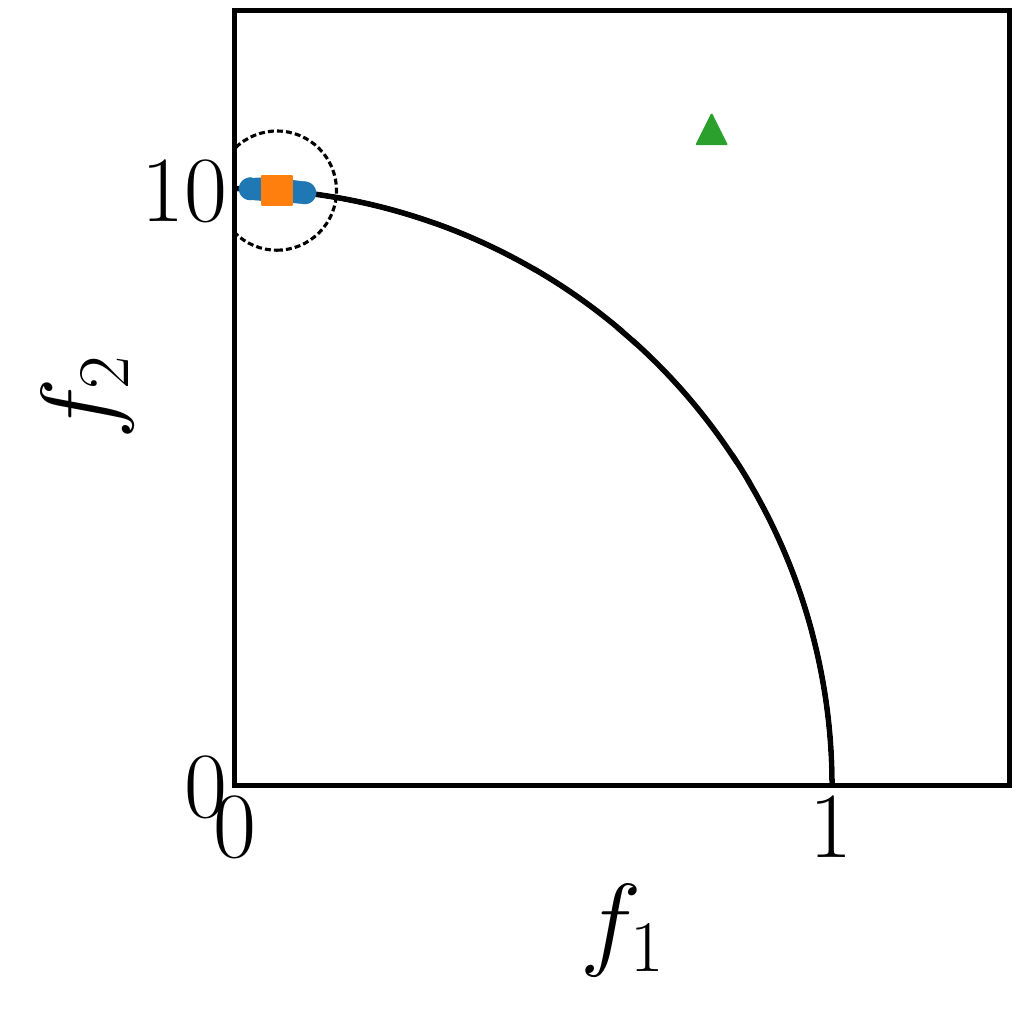}}
\subfloat[NR-NSGA-II]{\includegraphics[width=\wir\textwidth]{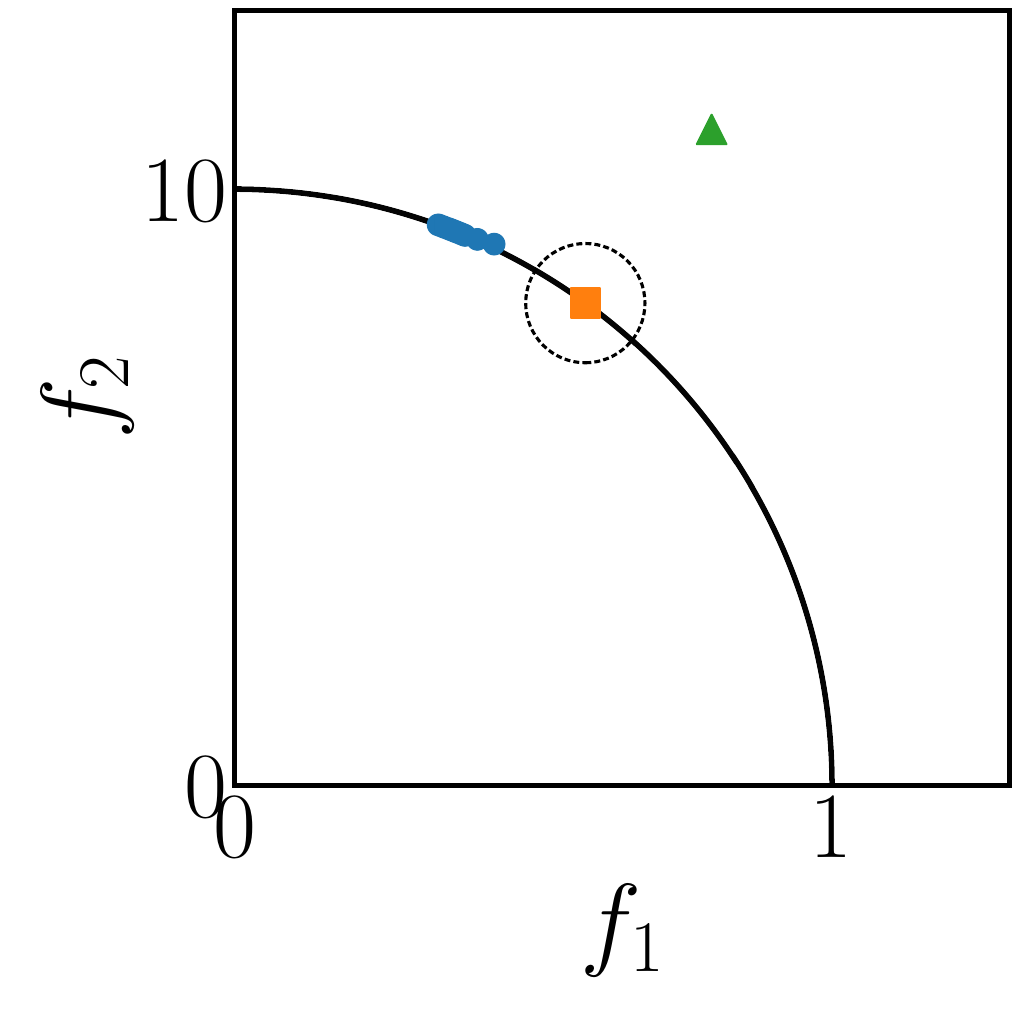}}
\caption{Distributions of the final populations of UR-NSGA-II and NR-NSGA-II on the SDTLZ2 problem. 
}
\label{fig:roic_sdtlz2}
\end{figure}

For further analysis, Figs. \ref{fig:nrnsga2_dtlz2_progress}(a)--(c) show the distributions of the populations obtained by NR-NSGA-II at $1\,000$, $3\,000$, and $5\,000$ function evaluations (\texttt{FE}), respectively.
\pref{fig:nrnsga2_dtlz2_progress} shows the result of the median-IGD$^+$-C run among the 31 runs.
In \pref{fig:nrnsga2_dtlz2_progress}, the symbol {\color{nearpt}{\rule{5pt}{5pt}}} represents the center point of the true UROI-C and NROI-C.
The symbol {\color{anearpt}\rotatebox{45}{\rule{5pt}{5pt}}} represents the approximation of {\color{nearpt}{\rule{5pt}{5pt}}} obtained from the current population $\set{P}$ of NR-NSGA-II.
More precisely, first, a set of non-dominated individuals $\set{R}$ is selected from $\set{P}$.
Then, the closest individual {\color{anearpt}\rotatebox{45}{\rule{5pt}{5pt}}} to the reference point $\vec{z}$ in the objective space is selected from $\set{R}$.
The semi-axis lengths $r_1$ and $r_2$ of the ellipsoid centered at {\color{anearpt}\rotatebox{45}{\rule{5pt}{5pt}}} are identical to those of the true UROI-C and NROI-C, i.e., $r_1 = 0.1$ and $r_2=0.1$. 
Note that, in this case, $r_1$ and $r_2$ are defined in the normalized objective space. 
Thus, the ellipsoid centered at {\color{anearpt}\rotatebox{45}{\rule{5pt}{5pt}}} represents the approximation of the NROI-C obtained from the current population of NR-NSGA-II.

In addition to \pref{fig:nrnsga2_dtlz2_progress}, \pref{fig:nrnsga2_dtlz2_progress_normalized} shows the distributions of the populations in NR-NSGA-II in the \textit{normalized} objective space.
Here, objective normalization in \pref{fig:nrnsga2_dtlz2_progress_normalized} is performed using \pref{eq:normalization_2}, not \pref{eq:normalization}.
The approximated ideal and nadir points ($\vec{z}^\text{lb}$ and $\vec{z}^\text{ub}$) are obtained from the above-mentioned set of non-dominated individuals $\set{R}$ as follows: $z^\text{lb}_i = \min_{\vec{x} \in \set{R}} f_i(\vec{x})$ and $z^\text{ub}_i = \max_{\vec{x} \in \set{R}} f_i(\vec{x})$ for $i \in \{1, \dots, m\}$.
Note that, as shown in \pref{fig:nrnsga2_dtlz2_progress}, the quality of $\vec{z}^\text{lb}$ ({\color{ideal}$\blacklozenge$}) and $\vec{z}^\text{ub}$ ({\color{nadir}$\blacklozenge$}) as approximations of the ideal and nadir points is poor, since PBEMO algorithms, including R-NSGA-II, do not approximate the whole PF~\cite{Tanabe24}.

As shown in \pref{fig:nrnsga2_dtlz2_progress}(a), because the individuals in the population are widely distributed in the objective space at the early stage of the search, the approximated region centered at {\color{anearpt}\rotatebox{45}{\rule{5pt}{5pt}}} is larger than the true NROI-C.
In this case, as shown in \pref{fig:nrnsga2_dtlz2_progress_normalized}(a), {\color{anearpt}\rotatebox{45}{\rule{5pt}{5pt}}} is close to {\color{nearpt}{\rule{5pt}{5pt}}}. 
As shown in \pref{fig:nrnsga2_dtlz2_progress}(b), the population of NR-NSGA-II has converged well to the PF at $3\,000$ function evaluations.
However, the approximated NROI-C is far from the true one, and its size is also smaller than that of the true NROI-C.
This is due to the poor quality of $\vec{z}^\text{lb}$ and $\vec{z}^\text{ub}$.
As seen in \pref{fig:nrnsga2_dtlz2_progress}(b), $\vec{z}^\text{lb}$ and $\vec{z}^\text{ub}$ are far from the true ideal and nadir points, respectively.
When objective normalization is performed using these $\vec{z}^\text{lb}$ and $\vec{z}^\text{ub}$, the distribution of the normalized objective vectors becomes biased.
In fact, as shown in \pref{fig:nrnsga2_dtlz2_progress_normalized}(b), a point far from the true center point {\color{nearpt}{\rule{5pt}{5pt}}} is nevertheless selected as a good approximation of it.
As shown in \pref{fig:nrnsga2_dtlz2_progress}(c) and \pref{fig:nrnsga2_dtlz2_progress_normalized}(c), the situation further deteriorates as the search progresses.
As demonstrated above, NR-NSGA-II fails to approximate the NROI-C even on problems with equally scaled objectives because the ideal and nadir points are poorly approximated.

\begin{tcolorbox}[sharpish corners, top=2pt, bottom=2pt, left=4pt, right=4pt, boxrule=0.0pt, colback=black!5!white,leftrule=0.75mm,]
\textbf{Summary and discussion}:
We demonstrated that NR-NSGA-II fails to approximate the NROI-C on the DTLZ problems with equally scaled objectives because of poor approximations of the ideal and nadir points ($\vec{z}^\text{lb}$ and $\vec{z}^\text{ub}$). 
The poor quality of $\vec{z}^\text{lb}$ and $\vec{z}^\text{ub}$ leads to a biased population distribution even on problems with equally scaled objectives, as shown in \pref{fig:roic_dtlz2}(b).
These observations suggest that the NROI-C is highly difficult to approximate, and the same holds for the NROI-A.
Approximating the NROI-C and NROI-A is impossible without the true ideal and nadir points ($\vec{z}^\text{ideal}$ and $\vec{z}^\text{nadir}$).
However, PBEMO algorithms can obtain only poor approximations of $\vec{z}^\text{ideal}$ and $\vec{z}^\text{nadir}$ because they do not approximate the whole PF.
This limitation is inevitable for all PBEMO algorithms.
\end{tcolorbox}

\begin{figure}[t]
\newcommand{\wir}{0.32}
\centering
\subfloat[$1\,000$ \texttt{FE}]{\includegraphics[width=\wir\textwidth]{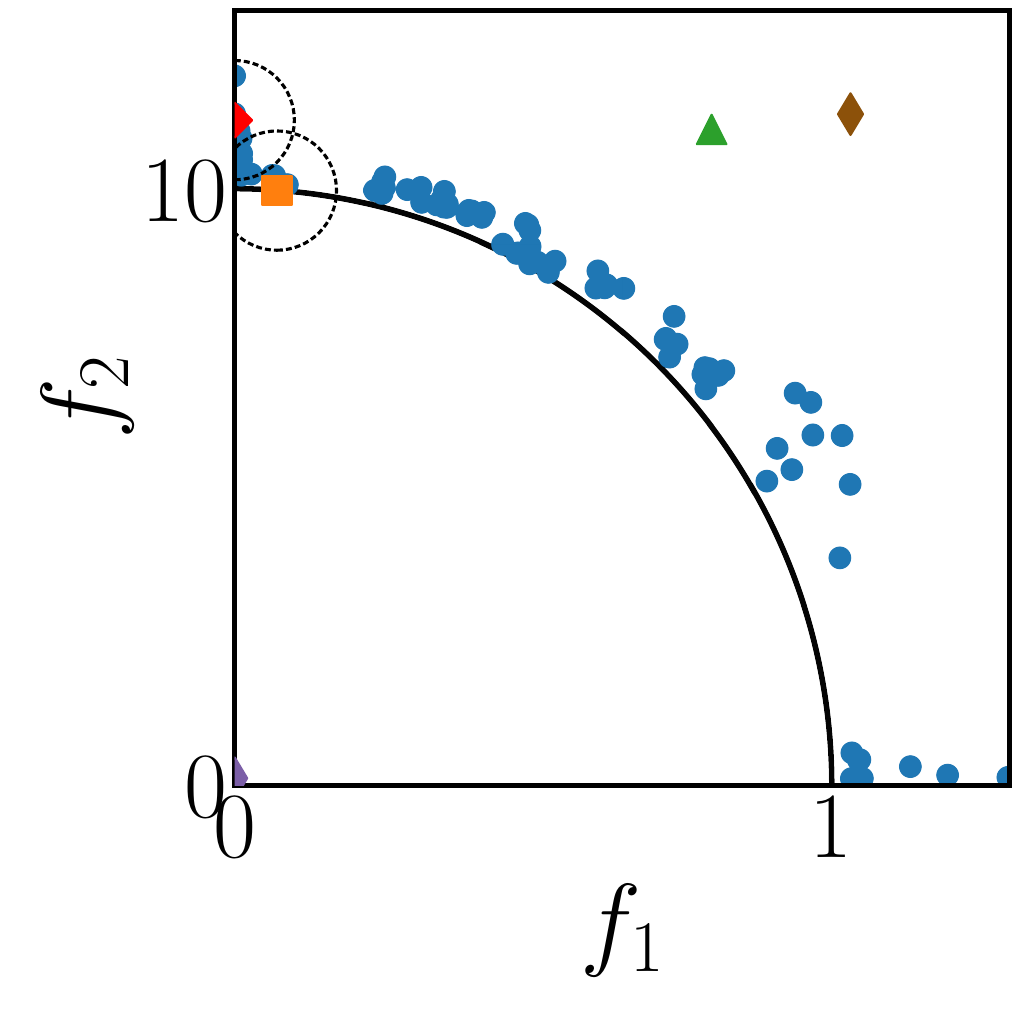}}
\subfloat[$3\,000$ \texttt{FE}]{\includegraphics[width=\wir\textwidth]{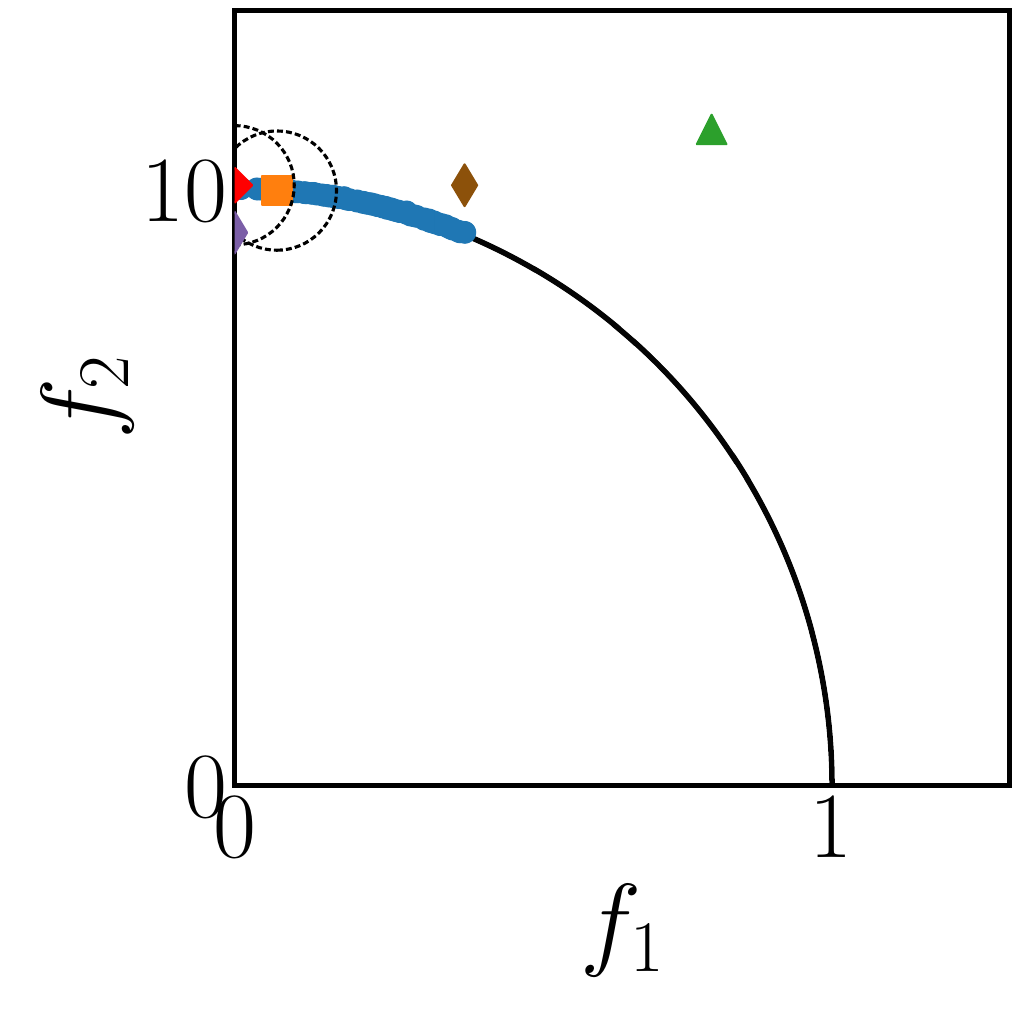}}
\subfloat[$5\,000$ \texttt{FE}]{\includegraphics[width=\wir\textwidth]{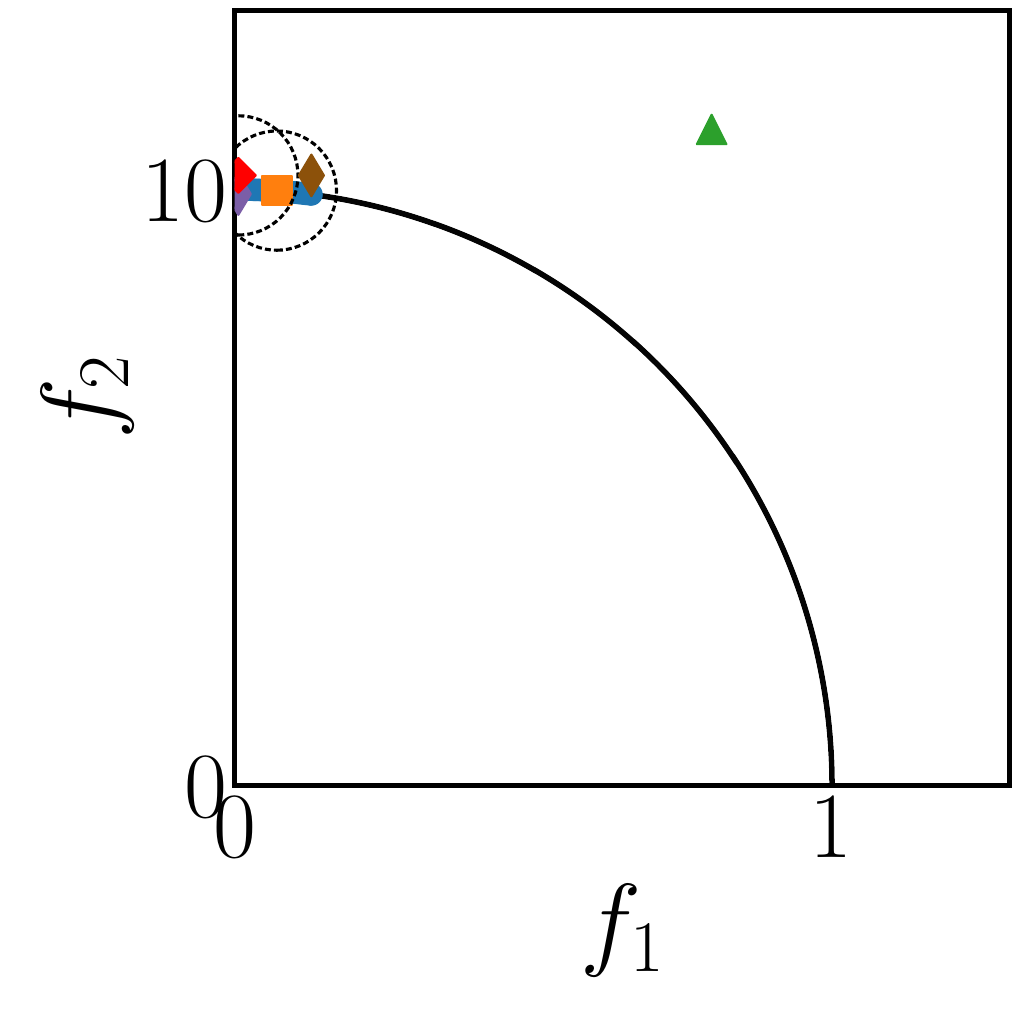}}\\
\caption{Distributions of the populations in UR-NSGA-II on the SDTLZ2 problem in the \textit{unnormalized} objective space.
}
\label{fig:urnsga2_sdtlz2_progress}
\end{figure}

\subsection{Why can UR-NSGA-II find a good approximation of the UROI-C on the SDTLZ problems?}
\label{subsec:discussion_sdtlz}

\pref{fig:roic_sdtlz2} shows the distributions of the final populations obtained by UR-NSGA-II and NR-NSGA-II on the bi-objective SDTLZ2 problem.
Similar to \pref{fig:roica_demo_sdtlz}(b), \pref{fig:roic_sdtlz2}(b) shows the results mapped onto the unnormalized objective space.

As shown in \pref{fig:roic_sdtlz2}(a), the final population obtained by UR-NSGA-II approximates the UROI-C well. 
As shown in \pref{tab:roic_comparison}(b), NR-NSGA-II achieves better $\text{IGD}^+\text{-C}$ values than UR-NSGA-II in approximating the NROI-C on the SDTLZ1--4 problems.
However, as shown in \pref{fig:roic_sdtlz2}(b), NR-NSGA-II fails to approximate the NROI-C. 
Here, the reason for the poor approximation obtained by NR-NSGA-II is the same as that explained in \pref{subsec:discussion_dtlz}.
Thus, the results of the relative comparison in \pref{tab:roic_comparison}(b) do not imply that NR-NSGA-II achieves a good approximation of the NROI-C in an absolute sense. 





As in \pref{fig:nrnsga2_dtlz2_progress}, Figs. \ref{fig:urnsga2_sdtlz2_progress}(a)--(c) show the population distributions obtained by UR-NSGA-II on the SDTLZ2 problem at $1\,000$, $3\,000$, and $5\,000$ function evaluations, respectively.
As shown in \pref{fig:urnsga2_sdtlz2_progress}, UR-NSGA-II, like NR-NSGA-II, yields poor approximations of the ideal and nadir points ($\vec{z}^\text{lb}$ and $\vec{z}^\text{ub}$).
However, UR-NSGA-II does not use any normalization mechanism, and the UROI-C to be approximated is defined in the original unnormalized objective space.
Thus, UR-NSGA-II is not affected by the poor quality of $\vec{z}^\text{lb}$ and $\vec{z}^\text{ub}$.






\begin{tcolorbox}[sharpish corners, top=2pt, bottom=2pt, left=4pt, right=4pt, boxrule=0.0pt, colback=black!5!white,leftrule=0.75mm,]
\textbf{Summary and discussion}:
We demonstrated that UR-NSGA-II can find a good approximation of the UROI-C despite the poor quality of $\vec{z}^\text{lb}$ and $\vec{z}^\text{ub}$, which hinders the approximation of the NROI-C, as discussed in \pref{subsec:discussion_dtlz}.
Fortunately, this issue does not arise when approximating the UROI-C, because PBEMO algorithms do not need to consider objective normalization.
In other words, approximating the UROI-C is much easier than approximating the NROI-C. The same holds for the UROI-A and UROI-P.


\end{tcolorbox}

%% file: conclusion.tex
\section{Conclusion}
\label{sec:conclusion}


This paper analyzed the effects of objective normalization on ROIs on the bi-objective DTLZ and SDTLZ problems. 
First, \pref{sec:roi_definition} demonstrated that the distributions of the UROI-C and NROI-C can differ significantly for problems with differently scaled objectives.
The same is true for the ROI-A, whereas the ROI-P is not affected by the choice of objective space.
Then, \pref{sec:results} investigated the performance of R-NSGA-II with and without objective normalization on the DTLZ and SDTLZ problems.
\pref{subsec:discussion_dtlz} demonstrated that the NROI-C and NROI-A are highly difficult to approximate because of poor approximations of the ideal and nadir points.
In contrast, \pref{subsec:discussion_sdtlz} showed that the UROI-C and UROI-A are much easier to approximate than the NROI-C and NROI-A.

Our findings are helpful for the DM in selecting the type of ROI.
The DM should clearly specify whether the ROI is defined in the unnormalized objective space or in the normalized objective space.
The DM should also keep in mind that good approximations of the NROI-C and NROI-A are highly unlikely to be obtained by PBEMO algorithms.
In contrast, good approximations of the UROI-C, UROI-A, UROI-P, and NROI-P can be obtained relatively easily by PBEMO algorithms.
%
Our findings are also helpful for practitioners conducting benchmarking.
As shown in \pref{tab:type_roi}, previous studies have not standardized either the type of ROI or whether the ROI is defined in the unnormalized or normalized objective space.
However, our findings suggest that the ROI should be defined explicitly to ensure reliable benchmarking.

Future work should investigate the effects of objective normalization on ROIs in real-world applications.
Our results suggest that approximating the NROI-C requires accurate estimates of the ideal and nadir points.
This issue may be addressed by incorporating Pareto corner search~\cite{SinghIR11} into PBEMO algorithms.
Further investigation of this direction is left for future work.


